\documentclass[10pt,twocolumn]{article}

\usepackage{cvpr}
\usepackage{times}
\usepackage{epsfig}
\usepackage{graphicx}
\usepackage{amsmath}
\usepackage{amssymb}

\usepackage{booktabs}
\usepackage{subcaption}
\usepackage{multirow}
\usepackage{xcolor}
\usepackage{tabularx}
\usepackage{colortbl}

\newcommand{\todo}[1]{}
\renewcommand{\todo}[1]{{\color{red} TODO: {#1}}}

\definecolor{DarkGreen}{rgb}{0.0, 0.5, 0.0}

\usepackage{color}
\definecolor{BrightRoyalPurple}{rgb}{0.65, 0.05, 0.78}

\newcommand{\comment}[1]{}

\usepackage[pagebackref=true,breaklinks=true,letterpaper=true,colorlinks,bookmarks=false]{hyperref}

\cvprfinalcopy % *** Uncomment this line for the final submission

\def\cvprPaperID{1914} % *** Enter the CVPR Paper ID here

\begin{document}

%%%%%%%%% TITLE

\title{3D Pose Estimation and 3D Model Retrieval for Objects in the Wild}

\author{Alexander Grabner$^1$\\
	\and
	Peter M. Roth$^1$\\
	\and
	Vincent Lepetit$^{2,1}$\\
	\and
	\small $^1$Institute of Computer Graphics and Vision, Graz University of Technology, Austria\\
	\and
	\small $^2$Laboratoire Bordelais de Recherche en Informatique, University of Bordeaux, France\\
	\and
	{\tt\small \{alexander.grabner,pmroth,lepetit\}@icg.tugraz.at}
}

\maketitle
%\thispagestyle{empty}

%%%%%%%%% ABSTRACT

\begin{abstract}

We propose a scalable, efficient and accurate approach to retrieve 3D models for objects in the wild. Our contribution is twofold. We first present a 3D pose estimation approach for object categories which significantly outperforms the state-of-the-art on Pascal3D+. Second, we use the estimated pose as a prior to retrieve 3D models which accurately represent the geometry of objects in RGB images. For this purpose, we render depth images from 3D models under our predicted pose and match learned image descriptors of RGB images against those of rendered depth images using a CNN-based multi-view metric learning approach. In this way, we are the first to report quantitative results for 3D model retrieval on Pascal3D+, where our method chooses the same models as human annotators for 50\% of the validation images on average. In addition, we show that our method, which was trained purely on Pascal3D+, retrieves rich and accurate 3D models from ShapeNet given RGB images of objects in the wild.

\end{abstract}

%%%%%%%%% BODY TEXT
\section{Introduction}

\begin{figure}
  
	\begin{subfigure}{\linewidth}
		\begin{center}
			%\fbox{\rule{0pt}{2in} \rule{\linewidth}{0pt}}
			\includegraphics[width=0.95\linewidth]{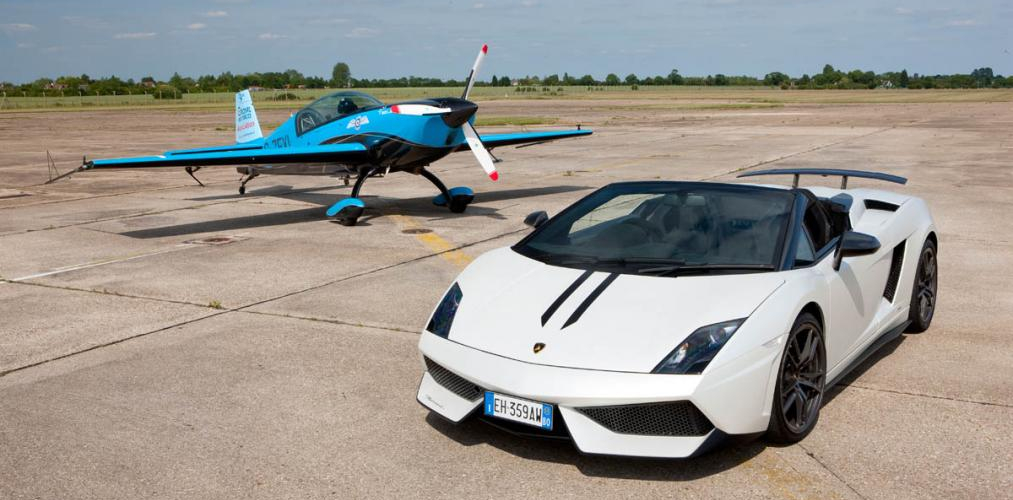}
			%\caption{}
		\end{center}
	\end{subfigure}\vspace{0.1cm}
	\begin{subfigure}{\linewidth}
		\begin{center}
			%\fbox{\rule{0pt}{2in} \rule{\linewidth}{0pt}}
			\includegraphics[width=0.95\linewidth]{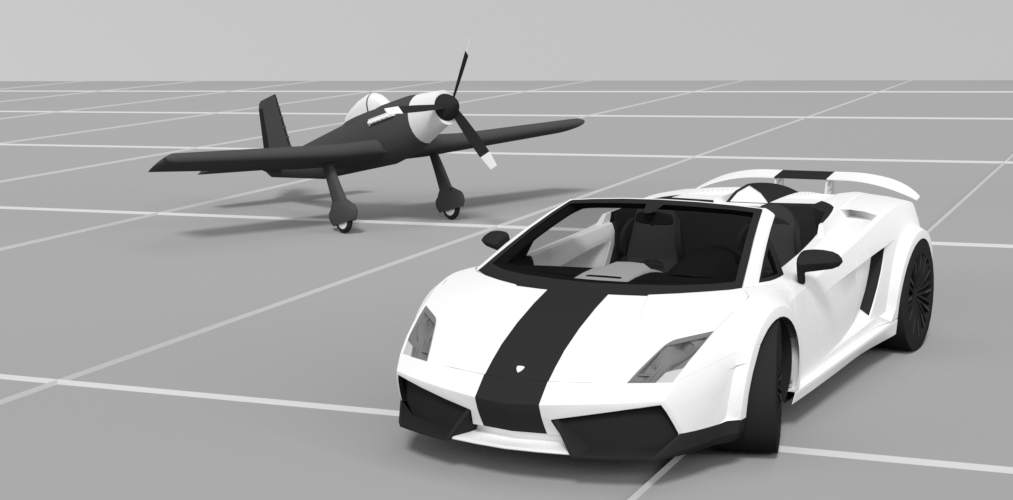}
			%\caption{}
		\end{center}
	\end{subfigure}
	\caption{Given an RGB image (top), we predict a 3D pose and a 3D model for objects of different categories (bottom).}
	\label{fig:teaser}
\end{figure}

Retrieving 3D models for objects in 2D images, as shown in Fig.~\ref{fig:teaser}, is extremely useful for 3D scene understanding, augmented reality applications and tasks like object grasping or object tracking. Recently, the emergence of large databases of 3D models such as ShapeNet~\cite{shapenet2015} initiated substantial interest in this topic and motivated research for matching 2D images of objects against 3D models. However, there is no straight forward approach to compare 2D images and 3D models, since they have considerably different representations and characteristics. 

One approach to address this problem is to project 3D models onto 2D images, which is known as rendering~\cite{su2015render}. This converts the task to comparing 2D images, which is, however, still challenging, because the appearance of objects in real images and synthetic renderings can significantly differ. In general, the geometry and texture of available 3D models do not exactly match those of objects in real images. Therefore, recent approaches~\cite{Aubry2015understanding,Izadinia2017im2cad,su2015multi,xiang2016objectnet3d} use convolutional neural networks (CNNs)~\cite{he2016deep,he2016identity,simonyan2014very} to extract features from images which are partly invariant to these variations. In particular, these methods compute image descriptors from real RGB images and synthetic RGB images which are generated by rendering 3D models under multiple poses. While this allows them to train a single CNN purely on synthetic data, there are two main disadvantages:

First, there is a significant domain gap between real and synthetic RGB images: Real images are affected by complex lighting, uncontrolled degradation and natural backgrounds. This makes it is hard to render photo-realistic images from the available 3D models. Therefore, using a single CNN for feature extraction from both domains is limited in performance, and even domain adaption~\cite{massa2016deep} does not fully account for the different characteristics of real and synthetic images.

Second, processing renderings from multiple poses is computationally expensive. However, this step is mandatory, because the appearance of an object can significantly vary with the pose, and mapping images from all poses to a common descriptor does not scale to many categories~\cite{li2015joint}. 

To overcome these limitations, we propose to first predict the object pose and to then use this pose as an effective prior for 3D model retrieval. Inspired by recent works on instance pose estimation~\cite{crivellaro2015novel,rad2017iccv}, we present a robust 3D pose estimation approach for object categories based on virtual control points. More specifically, we use a CNN to predict the 2D projections of virtual 3D control points from which we recover the pose using a P$n$P algorithm. This approach does not only outperform the state-of-the-art for viewpoint estimation on Pascal3D+~\cite{xiang2014beyond}, but also supports category-agnostic predictions. Having an estimate of the 3D pose makes our approach scalable, as it reduces the matching process to a single rendering per 3D model.

Additionally, we propose to render depth images instead of RGB images and to use different CNNs for feature extraction from the real and synthetic domain. Thus, we are not only able to deal with untextured models, but also to alleviate the domain gap. We implement our 3D model retrieval method using a multi-view metric learning approach, which is trained on real and synthetic data from Pascal3D+. In this way, we are the first to present quantitative results for 3D model retrieval on Pascal3D+. Moreover, we demonstrate that our approach retrieves rich and accurate 3D models from ShapeNet given unseen images from Pascal3D+. To summarize, we make the following contributions:
\begin{itemize}
	\item[--] We present a 3D pose estimation approach for object categories which significantly outperforms the state-of-the-art on Pascal3D+. Our method predicts virtual control points which generalize across categories making the approach scalable.

	\item[--] We introduce a 3D model retrieval approach which utilizes a pose prior. For this purpose, we match learned image descriptors of RGB images against those of depth images rendered from 3D models under our predicted pose. In this way, we retrieve 3D models from ShapeNet which accurately represent the geometry of objects in RGB images, as shown in Fig.~\ref{fig:teaser}.
\end{itemize}

\section{Related Work}

Since there is a vast amount of literature on both 3D pose estimation and 3D model retrieval, we focus our discussion on recent works which target these tasks for object categories in particular.

\subsection{3D Pose Estimation}

Many recent works only perform 3-DoF viewpoint estimation and predict the object rotation using regression, classification or hybrid variants of the two. \cite{xiang2016objectnet3d} directly regresses azimuth, elevation and in-plane rotation using a CNN. \cite{massa2016crafting} compares different variants and presents a regression approach which parameterizes each angle using trigonometric functions. \cite{tulsiani2015pose,tulsiani2015viewpoints} perform viewpoint classification by discretizing the range of each angle into a number of disjoint bins and predicting the most likely bin using a CNN. \cite{su2015render} uses a fine-grained geometric structure aware classification, which encourages the correlation between bins of nearby views. \cite{mousavian20163d} formulates the task as a hybrid classification/regression problem: In addition to viewpoint classification, a residual rotation is regressed for each angular bin, and the 3D dimensions of the object are predicted. \cite{mottaghi2015coarse} uses a slightly different parameterization and predicts a 2D translation to refine \mbox{the object localization in a coarse-to-fine hybrid approach.}

However, predicting a full \hbox{6-DoF} pose instead of a \hbox{3-DoF} viewpoint is desirable for many applications. Therefore, numerous methods compute both rotation and translation from 2D/3D keypoint correspondences. \cite{pepik20153d} recovers the pose from keypoint predictions and CAD models using a P$n$P algorithm. \cite{tulsiani2015viewpoints} presents a keypoint prediction approach that combines local keypoint estimates with a global viewpoint estimate. \cite{pavlakos17object3d} predicts semantic keypoints and trains a deformable shape model which takes keypoint uncertainties into account.

These approaches rely on category-specific keypoints which do not generalize across categories. In the context of 3D pose estimation for object instances, \cite{crivellaro2015novel} therefore considers virtual control points and predicts their 2D projections to estimate the pose from object parts. \cite{rad2017iccv} takes a similar approach, but uses the corners of the object's 3D bounding box as virtual control points. This work inspired our approach, however, it is not directly applicable for object category pose estimation, since the ground truth 3D model of an object must be known at runtime.

\subsection{3D Model Retrieval}

One intuitive approach to 3D model retrieval is to rely on classification. \cite{mottaghi2015coarse} performs fine-grained category recognition and provides a model for each category. \cite{aubry2014seeing} uses a linear classifier on mid-level representations of real images and renderings from multiple viewpoints to predict both shape and viewpoint. 

However, retrieval via classification does not scale. Therefore, many recent methods take a metric learning approach. The most common strategy is to train a single CNN to extract features from real RGB images and RGB renderings. \cite{Aubry2015understanding} uses a CNN pre-trained on ImageNet~\cite{russakovsky2015imagenet} as a feature extractor and matches features of real images against those of 3D models rendered under multiple viewpoints to predict both shape and viewpoint. \cite{Izadinia2017im2cad} takes a similar approach, but uses a different network architecture for feature extraction. \cite{massa2016deep} also employs a pre-trained CNN, but additionally performs non-linear feature adaption to overcome the domain gap between real and rendered images.

\cite{xiang2016objectnet3d} finetunes a pre-trained CNN using lifted structure embedding~\cite{oh2016deep} and averages the distance of a real image to renderings from multiple viewpoints to be more invariant to object pose. \cite{su2015multi} presents a CNN architecture that combines information of renderings from multiple viewpoints into a single object pose invariant descriptor. \cite{li2015joint} explicitly constructs an embedding space using a 3D similarity measure evaluated on clean 3D models and trains a CNN to map renderings with arbitrary backgrounds to the corresponding points in the embedding space.

While it is convenient to use RGB images, it is unclear how to deal with untextured 3D models or how to set the scene lighting. Therefore, other methods perform 3D model retrieval using depth instead of RGB images. \cite{Feng20163d} uses an ensemble of autoencoders followed by a domain adaption layer to match real depth images against depth images of 3D models. \cite{Zhu2016deep} computes image descriptors by fusing global autoencoder and local SIFT features of depth images. However, real depth images are not available in many scenarios.

Another approach which alleviates the domain gap and maps different representations to a common space is multi-view learning. \cite{girdhar2016learning} trains two different networks to map 3D voxel grids and RGB images to a low dimensional embedding space, where 3D model retrieval is performed by matching embeddings of real RGB images against those of voxel grids. \cite{Zhu2015learning} also presents a multi-view approach using two networks, but maps LD-SIFT features extracted from 3D models and depth images to a common space. In contrast to these methods, we map real RGB images and rendered depth images to a common representation. In this way, we do not need to perform computationally expensive 3D convolutions for high-resolution voxel grids and do not rely on real depth images.

\section{3D Pose Estimation and 3D Model Retrieval}
\label{sec:method}

\begin{figure*}
	\begin{center}
		%\fbox{\rule{0pt}{2in} \rule{\linewidth}{0pt}}
		\includegraphics[width=.95 \linewidth]{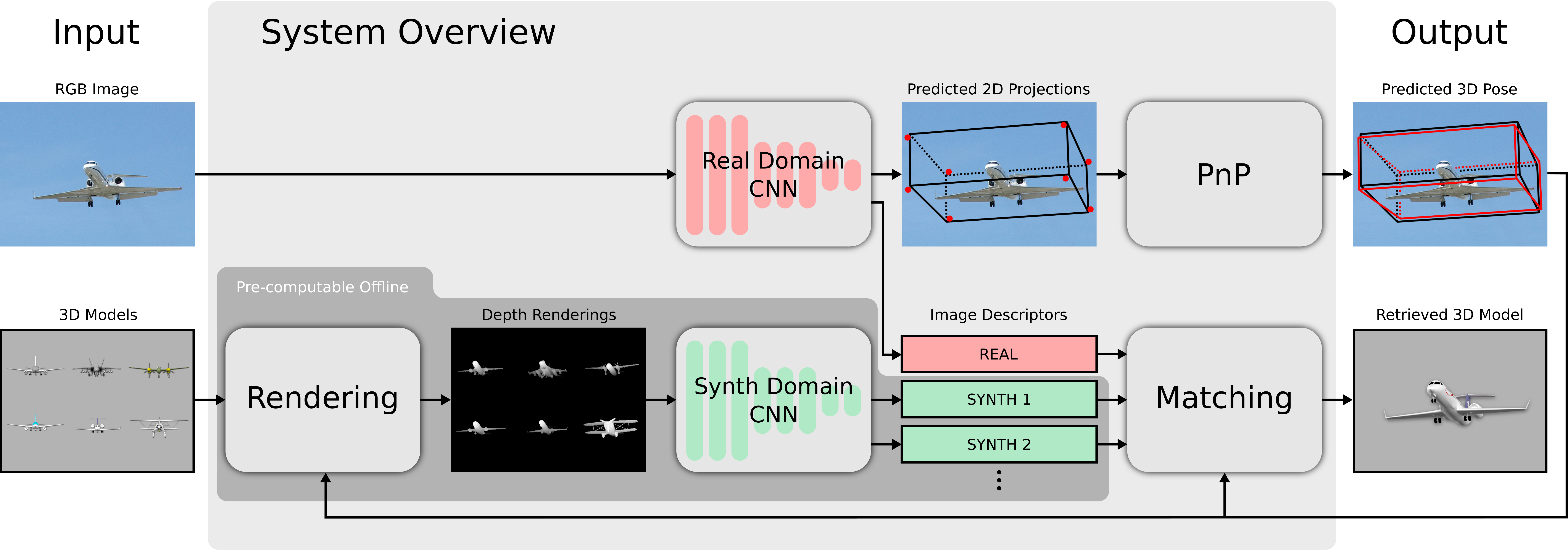}
		\caption{Overview of our approach. First row: Given an RGB image of an object, we first predict its 3D pose. We use a CNN to predict the 2D projections of the object's 3D bounding box corners (red dots). From these, we recover the object pose using a P$n$P algorithm. Second row: We render depth images from 3D models under the estimated pose and extract image descriptors from the real RGB image and the synthetic depth images using two different CNNs. Finally, we match the computed descriptors to retrieve the closest 3D model. Our approach supports pre-computed synthetic descriptors.}
	\label{fig:overview}
	\end{center}
    \vspace{-0.75cm}
\end{figure*}

Given an RGB image containing one or more objects, we want to retrieve 3D models with a geometry that corresponds well to the actual objects. Fig.~\ref{fig:overview} shows our proposed pipeline. We first estimate the 3D pose of an object from an image window roughly centered on the object.  In this work, we assume the input image windows are known as in~\cite{xiang2014beyond} or given by a 2D object detector~\cite{ren2015faster}. Similar to previous works~\cite{mousavian20163d,pavlakos17object3d,tulsiani2015viewpoints}, we also assume the object category to be known, as it is a useful prior for both pose estimation and model retrieval. However, we also show that this information is not necessarily required in our approach. In fact, we can retrieve an accurate pose with only a marginal loss of accuracy, when the category is unknown.

After we estimated the object pose, we render a number of candidate 3D models under that pose. In particular, we render depth images, which allows us to deal with untextured 3D models and to circumvent the problem of scene lighting. In order to compare the real RGB image to synthetic depth renderings, we extract image descriptors using two CNNs, one for each domain. Finally, we match these image descriptors to retrieve the closest 3D model.

\subsection{3D Pose Estimation}
\label{sec:3dpose}

\newcommand{\pose}{\text{pose}}
\newcommand{\proj}{\text{proj}}
\newcommand{\dime}{\text{dim}} % \dim is already a latex command
\newcommand{\reg}{\text{reg}}
\newcommand{\huber}{\text{Huber}}

The first step in our model retrieval approach is to robustly compute the 3D pose of the objects of interest. For this purpose, inspired by~\cite{crivellaro2015novel,rad2017iccv}, we predict the 2D image locations of virtual control points.  More precisely, we train a CNN to predict the 2D image locations of the projections of the object's eight 3D bounding box corners. The actual 3D pose is then computed by solving a perspective-$n$-point (P$n$P) problem, which recovers rotation and translation from 2D-3D correspondences.  This is illustrated in the first row of Fig.~\ref{fig:overview}.

However, P$n$P algorithms require the 3D coordinates of the virtual control points to be known. Therefore, previous approaches either assume the exact 3D model to be given at runtime~\cite{rad2017iccv} or predict the projections of static 3D points~\cite{crivellaro2015novel}. To overcome this limitation, we predict the spatial dimensions $D=[d_x,d_y,d_z]$ of the object's 3D bounding box and use these to scale a unit cube, which approximates the ground truth 3D coordinates.

For this purpose, we introduce a CNN architecture which jointly predicts the 2D image locations of the projections of the eight 3D bounding box corners (16 values) as well as the 3D bounding box dimensions (3 values). As illustrated in Fig.~\ref{fig:loss}, we implement this architecture as a single 19 neuron linear output layer, which we apply on top of the penultimate layer of different base networks such as VGG~\cite{simonyan2014very} or ResNet~\cite{he2016deep,he2016identity}.
During training, we optimize the pose loss 
\begin{equation}
L_\pose = L_\proj + \alpha L_\dime + \beta L_\reg  \> ,
\label{eq:pose_loss}
\end{equation}

\noindent which is a linear combination of the projection loss $L_\proj$, the dimension loss $L_\dime$ and the regularization $L_\reg$. The meta-parameters $\alpha$ and $\beta$ control the impact of the different loss terms. 
Let $\text{M}_i$ be the $i$-th 3D bounding box corner and $\text{Proj}_{\text{R,t}}(\text{M}_i)$ its projection using the ground truth rotation R and translation t, then the projection loss
%
% \alexrmk{Should we know use the expected value or not??}
%
\begin{equation}
L_\proj = E\left[ \sum_{i=1}^8 \Vert \text{Proj}_{\text{R,t}}(\text{M}_i) - \widetilde{m}_i\Vert_\huber \right]
\end{equation} 
\noindent is the expected value of the distances between the ground truth projections $\text{Proj}_{\text{R,t}}(\text{M}_i)$ and the predicted locations of these projections $\widetilde{m}_i$ computed by the CNN for the training set. Being aware of inaccurate annotations in datasets such as Pascal3D+~\cite{xiang2014beyond}, we use the Huber loss~\cite{huber1964robust} in favor of the squared loss to be more robust to outliers. 

The dimension loss 
\begin{equation}
L_\dime = E\left[ \sum_{i=x,y,z} \Vert d_i - \widetilde{d}_i\Vert_\huber \right]
\end{equation} 
\noindent is the expected value of the distances between the ground truth 3D dimensions $d_i$ and the  3D dimensions $\widetilde{d}_i$ predicted by the CNN for the training set. To reduce the risk of overfitting, the regularization $L_\reg$ in Eq.~(\ref{eq:pose_loss}) adds weight decay for all CNN parameters.

\subsection{3D Model Retrieval}
\label{sec:3dretrieval}

Having a robust estimate of the object pose, we render 3D models under this pose instead of rendering them under multiple poses~\cite{Aubry2015understanding,Izadinia2017im2cad,massa2016deep,su2015multi}. This significantly reduces the computational complexity compared to methods which process multiple renderings for each 3D model and provides a useful prior for retrieval. In contrast to recent approaches~\cite{li2015joint,massa2016deep,su2015multi,xiang2016objectnet3d}, we render depth images instead of RGB images. This allows us to deal with 3D models which do not have material or texture. Additionally, we circumvent the problem of how to set the scene lighting.

Before rendering a 3D model, we re-scale it to tightly fit into our predicted 3D bounding box. This is done by multiplying all vertices with the minimum of the ratio between the predicted 3D dimensions computed during pose estimation and the model's actual 3D dimensions. In this way, we improve the alignment between input RGB images and rendered depth images.

\begin{figure}
	\begin{center}
		%\fbox{\rule{0pt}{2in} \rule{\linewidth}{0pt}}
		\includegraphics[width=\linewidth]{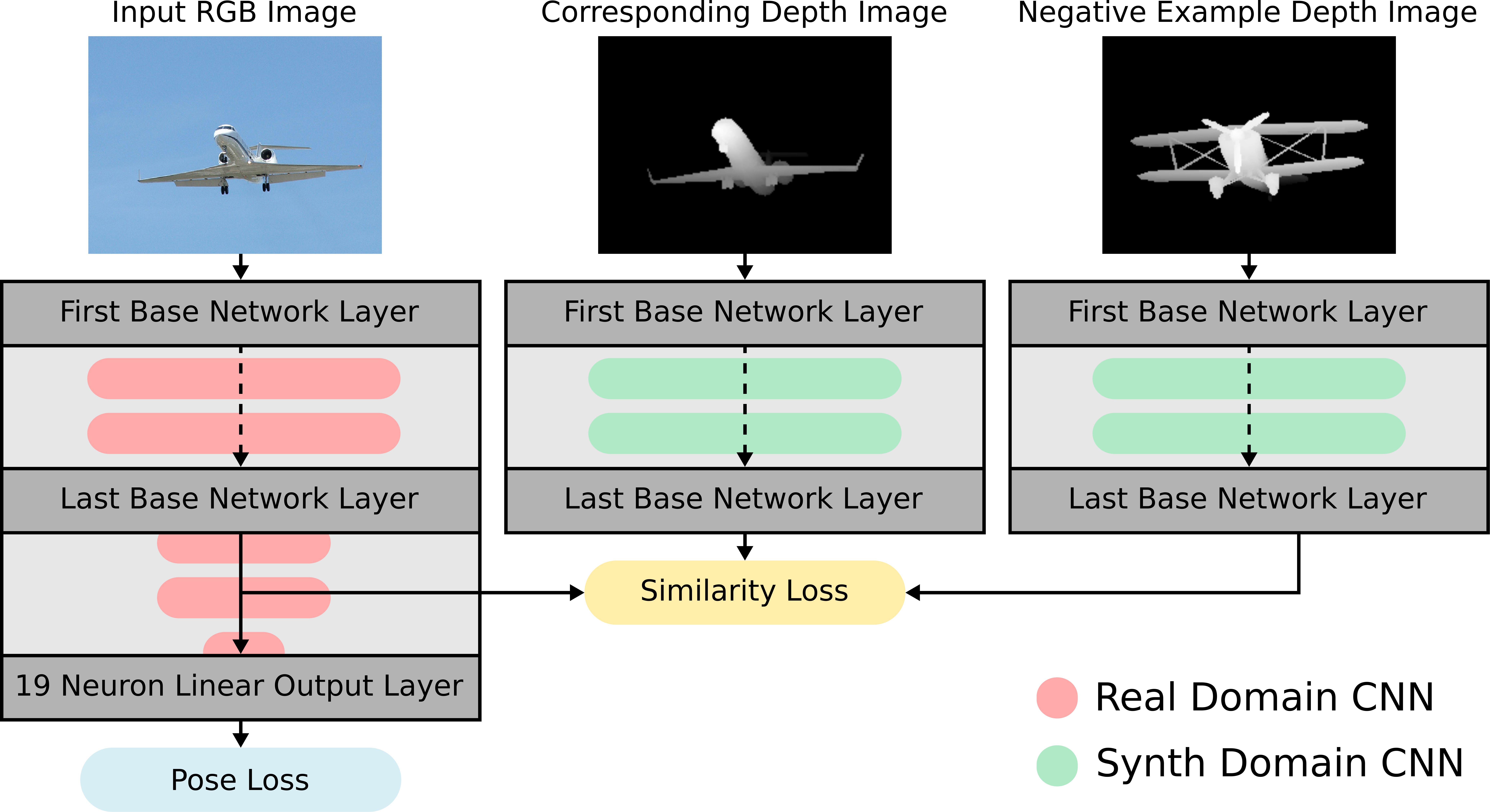}
		\caption{The pose loss is computed on the output of the real domain CNN. The similarity loss is computed on hidden feature maps extracted from the last base network layer of the real and synthetic domain CNN.}
	\label{fig:loss}
	\end{center}
    \vspace{-0.75cm}
\end{figure}

However, since RGB images and depth images have considerably different characteristics, we introduce a multi-view metric learning approach, which maps images from both domains to a common representation. We implement this mapping using a separate CNN for each domain.
For real RGB images, we extract image descriptors from the hidden feature activations of the penultimate layer of our pose estimation CNN (see Fig.~\ref{fig:loss}). As these activations have already been computed during pose estimation inference, we get the real image descriptor without any additional computational cost.
For the synthetic depth images, we extract image descriptors using a CNN with the same architecture as our pose estimation CNN, except for the output layer (see Fig.~\ref{fig:loss}).

To finally map images from both domains to a common representation, we optimize the similarity loss
%
% \vspace{-0.10cm}
\begin{equation}
L_\text{similarity} = L_\text{descr} + \gamma L_{\text{reg}_2} \> ,
\label{eq:similarity_loss}
\end{equation}
\noindent which comprises the image descriptor loss $L_\text{descr}$ and the regularization $L_{\text{reg}_2}$ weighted by the meta-parameter $\gamma$.

The image descriptor loss
\begin{equation}
L_\text{descr} =  E\left[ \max(0,s^+ - s^- + m) \right]
\end{equation}
\noindent minimizes the expected value of the Triplet loss~\cite{weinberger2009distance} for the training set. Here, $s^+$ is the Euclidean distance between the real RGB image descriptor and the corresponding synthetic depth image descriptor, $s^-$ is the Euclidean distance between the real RGB image descriptor and a negative example synthetic depth image descriptor, and $m$ specifies the margin, \ie, the desired minimum difference between $s^+$ and $s^-$. To reduce the risk of overfitting, the regularization $L_{\text{reg}_2}$ in Eq.~(\ref{eq:similarity_loss}) adds weight decay for all CNN parameters.

After the optimization of the CNNs, we can pre-compute descriptors for synthetic depth images. In this case, we generate multiple renderings for each 3D model, which cover the full pose space. We then compute descriptors for all these renderings and store them in a database. At runtime, we just match descriptors from the viewpoint closest to our predicted pose, which is fast and scalable, but still accurate as shown in our experiments.

\section{Experimental Results}

\newcommand{\gt}{\text{gt}}
\newcommand{\pred}{\text{pred}}

To demonstrate our 3D model retrieval approach for objects in the wild, we evaluate
it in a realistic setup where we retrieve 3D models from ShapeNet~\cite{shapenet2015} given unseen RGB images from Pascal3D+~\cite{xiang2014beyond}. In particular, we train our 3D model retrieval approach purely on data from Pascal3D+, but use it to retrieve 3D models from ShapeNet. The corresponding results are detailed in Sec.~\ref{sec:exp-retrieval}. As estimating an accurate object pose is essential for our retrieval approach, we additionally evaluate our pose estimation approach on Pascal3D+ in Sec.~\ref{sec:exp-pose}.

\subsection{3D Pose Estimation}
\label{sec:exp-pose}

In the following, we first give a detailed evaluation of our pose estimation approach. Then, we compare it to previous methods, outperforming the state-of-the-art for viewpoint estimation on Pascal3D+. Finally, we demonstrate that we are even able to top the state-of-the-art without providing the correct category prior in some cases. For a fair evaluation, we follow the evaluation protocol of \cite{tulsiani2015viewpoints}, which quantifies 3-DoF viewpoint prediction accuracy on Pascal3D+ using the geodesic distance
\begin{equation}
\Delta(R_\gt,R_\pred) = \frac{\Vert \text{log}(R_\gt^T R_\pred^{\vphantom{T}} )\Vert_F}{\sqrt{2}}
\end{equation}
\noindent to measure the difference between the ground truth viewpoint rotation matrix $R_\gt$ and the predicted viewpoint rotation matrix $R_\pred$. In particular, we report two metrics: $MedErr$ (the median of all viewpoint differences) and $Acc_{\frac{\pi}{6}}$ (the percentage of all viewpoint differences smaller than $\frac{\pi}{6}$ respectively $30^\circ$). Evaluating our approach using the $AVP$ metric~\cite{xiang2014beyond}, which couples 2D object detection and azimuth classification, is not meaningful as it is very different from our specific task.

\subsubsection{3D Pose Estimation on Pascal3D+}

Table~\ref{table:viewpoint-variants} presents quantitative results for 3-DoF viewpoint estimation on Pascal3D+ using our approach in different setups, starting from a baseline using VGG to a more elaborated version building on ResNet. Specific implementation details and other parameters are provided in the supplementary material. For our baseline approach~(\emph{Ours - VGG}) we build on VGG and fine-tune the entire network for our task similar to \cite{mousavian20163d,tulsiani2015pose,tulsiani2015viewpoints}. As can be seen from Table~\ref{table:viewpoint-sota}, this baseline already matches the state-of-the-art.

\begin{table}[t]
	\centering
	\begin{tabular}{lcc}
		\toprule
		&\multicolumn{1}{c}{$MedErr$}&\multicolumn{1}{c}{$Acc_{\frac{\pi}{6}}$}\\
		\midrule
		Ours - VGG&11.7&0.8076\\
		Ours - VGG+blur&11.6&0.8033\\
		Ours - ResNet&\bfseries10.9&0.8341\\
		Ours - ResNet+blur&\bfseries10.9&\bfseries0.8392\\
		\bottomrule
	\end{tabular}
	\caption{Viewpoint estimation using ground truth detections on Pascal3D+ for different setups of our approach. We report the mean performance across all categories.}
	\label{table:viewpoint-variants}
	\vspace{-0.25cm}
\end{table}

When inspecting the failure cases, we see that many of them relate to small objects. In these cases, object image windows need to be upscaled to fit the fixed spatial input resolution of pre-trained CNNs. This results in blurry images and VGG, which only employs  3$\times$3 convolutions, performs poorly at extracting features from over-smoothed images.

Therefore, we propose to use a network with larger kernel sizes that performs better at handling over-smoothed input images such as ResNet50~\cite{he2016deep,he2016identity}, which uses 7$\times$7 kernels in the first convolutional layer. As presented in Table~\ref{table:viewpoint-variants}, our approach with ResNet-backend (\emph{Ours - ResNet}) significantly outperforms the VGG-based version. In addition, the total number of network parameters is notably lower (VGG: 135M vs. ResNet: 24M).

To further improve the performance, we employ data augmentation in the form of image blurring. Using ResNet as a base network together with blurring training images (\emph{Ours ResNet+blur}), we improve on the $Acc_{\frac{\pi}{6}}$ metric while maintaining low $MedErr$ (see Table~\ref{table:viewpoint-variants}). This indicates that we improve the performance on over-smoothed images, but do not loose accuracy on sharp images. While our approach with ResNet-backend shows increased performance in this setup, we do not benefit from training on blurred images using a VGG-backend (\emph{Ours - VGG+blur}). This also confirms that VGG is not suited for feature extraction from over-smoothed images. For all following experiments, we use our best performing setup, \ie, \emph{Ours - ResNet+blur}.

\definecolor{lightgreen}{RGB}{245,255,245}
\begin{table*}[h]
	\centering
	\begin{tabular}{lrrrrrrrrrrrr|X>{\columncolor{lightgreen}}c}
		\toprule
		&\multicolumn{13}{c}{\bfseries category-specific}\\
		\cmidrule{2-14}
		&\multicolumn{1}{c}{aero}&\multicolumn{1}{c}{bike}&\multicolumn{1}{c}{boat}&\multicolumn{1}{c}{bottle}&\multicolumn{1}{c}{bus}&\multicolumn{1}{c}{car}&\multicolumn{1}{c}{chair}&\multicolumn{1}{c}{table}&\multicolumn{1}{c}{mbike}&\multicolumn{1}{c}{sofa}&\multicolumn{1}{c}{train}&\multicolumn{1}{c|}{tv}&\multicolumn{1}{c}{\cellcolor{lightgreen}mean}\\
		\midrule
		$MedErr$ (\cite{pavlakos17object3d})&11.2&15.2&37.9&13.1&4.7&6.9&12.7&N/A&N/A&21.7&9.1&38.5&N/A\\
		$MedErr$ (\cite{pavlakos17object3d}*)&\bfseries8.0&13.4&40.7&11.7&\bfseries2.0&5.5&10.4&N/A&N/A&9.6&8.3&32.9&N/A\\
		$MedErr$ (\cite{tulsiani2015viewpoints})&13.8&17.7&21.3&12.9&5.8&9.1&14.8&15.2&14.7&13.7&8.7&15.4&13.6\\
		$MedErr$ (\cite{mousavian20163d})&13.6&12.5&22.8&\bfseries8.3&3.1&5.8&11.9&12.5&12.3&12.8&6.3&11.9&11.1~\\
		$MedErr$ (\cite{su2015render}**)&15.4&14.8&25.6&9.3&3.6&6.0&\bfseries9.7&\bfseries10.8&16.7&\bfseries9.5&\bf6.1&12.6&11.7\\
		$MedErr$ (Ours) &10.0&15.6&\bfseries19.1&8.6&3.3&\bfseries5.1&13.7&11.8&\bfseries12.2&13.5&6.7&\bfseries11.0&\bfseries10.9\\
		\midrule
		$Acc_{\frac{\pi}{6}}$ (\cite{tulsiani2015viewpoints}) & 0.81 & 0.77 & 0.59 & 0.93   &\bfseries0.98 & 0.89 & 0.80  & 0.62  &0.88  & 0.82 &0.80  & 0.80 & 0.8075 \\
		$Acc_{\frac{\pi}{6}}$ (\cite{mousavian20163d}) & 0.78 &\bfseries0.83 & 0.57 & 0.93   & 0.94 & 0.90 &0.80  &0.68  & 0.86  & 0.82 &0.82  & 0.85 & 0.8103 \\
		$Acc_{\frac{\pi}{6}}$ (\cite{su2015render}**) & 0.74 &\bfseries0.83 & 0.52 & 0.91 & 0.91 &0.88 &\bfseries0.86 &\bfseries0.73 &0.78&\bfseries0.90&\bfseries0.86&\bfseries0.92&0.8200 \\
		$Acc_{\frac{\pi}{6}}$ (Ours)&\bfseries0.83&0.82&\bfseries0.64&\bfseries0.95&0.97&\bfseries0.94&0.80&0.71&0.88&0.87&0.80&0.86&\bfseries0.8392\\
		\midrule[\heavyrulewidth]
		&\multicolumn{13}{c}{\bfseries category-agnostic}\\
		\cmidrule{2-14}
		&\multicolumn{1}{c}{aero}&\multicolumn{1}{c}{bike}&\multicolumn{1}{c}{boat}&\multicolumn{1}{c}{bottle}&\multicolumn{1}{c}{bus}&\multicolumn{1}{c}{car}&\multicolumn{1}{c}{chair}&\multicolumn{1}{c}{table}&\multicolumn{1}{c}{mbike}&\multicolumn{1}{c}{sofa}&\multicolumn{1}{c}{train}&\multicolumn{1}{c|}{tv}&\multicolumn{1}{c}{\cellcolor{lightgreen}mean}\\
		\midrule
		$MedErr$ (Ours)&10.9&\bfseries12.2&23.4&9.3&3.4&5.2&15.9&16.2&\bfseries12.2&11.6&6.3&11.2&11.5\\
		\midrule
		$Acc_{\frac{\pi}{6}}$    (Ours)&0.80&0.82&0.57&0.90&0.97&\bfseries0.94&0.72&0.67&\bfseries0.90&0.80&0.82&0.85&0.8133\\
		\bottomrule
	\end{tabular}
	\caption{Viewpoint estimation using ground truth detections on Pascal3D+. * The ground truth 3D model must be known at runtime. ** The approach was trained on vast amounts of RGB renderings from ShapeNet, instead of Pascal3D+ data.}
	\label{table:viewpoint-sota}
	\vspace{-0.35cm}
\end{table*}

\subsubsection{Comparison to the State-of-the-Art}

Next, we compare our pose estimation approach to state-of-the-art methods on Pascal3D+. Quantitative results are presented in Table~\ref{table:viewpoint-sota}. Our approach significantly outperforms the state-of-the-art in both $MedErr$ and $Acc_{\frac{\pi}{6}}$ considering mean performance across all categories and also shows competitive results for individual categories.

However, the $Acc_{\frac{\pi}{6}}$ scores for two categories, \textit{boat} and \textit{table}, are significantly below the mean. We analyze these results in more detail. The category \textit{boat} is the most challenging category due to the large intra-class variability in shape and appearance. Many detections for this category are of low resolution and often objects are barely visible because of fog or mist. Additionally, there are a lot of ambiguities, \eg, even a human cannot distinguish between the front and the back of an unmanned canoe. Nevertheless, we outperform the state-of-the-art for this challenging category.

\begin{figure*}[!h]
  \begin{subfigure}{0.095\linewidth}
    \begin{center}
      %\fbox{\rule{0pt}{2in} \rule{\linewidth}{0pt}}
      \includegraphics[width=\linewidth]{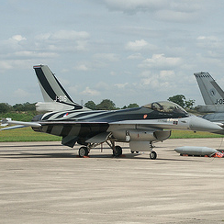}
      %\caption{}
    \end{center}
  \end{subfigure}\hfill\begin{subfigure}{0.095\linewidth}
  \begin{center}
    %\fbox{\rule{0pt}{2in} \rule{\linewidth}{0pt}}
    \includegraphics[width=\linewidth]{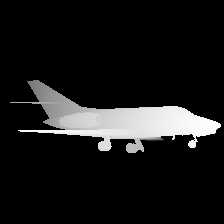}
    %\caption{}
  \end{center}
  \end{subfigure}\hfill\begin{subfigure}{0.095\linewidth}
  \begin{center}
    %\fbox{\rule{0pt}{2in} \rule{\linewidth}{0pt}}
    \includegraphics[width=\linewidth]{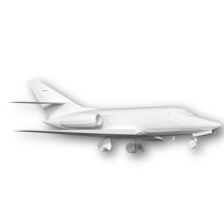}
    %\caption{}
  \end{center}
  \end{subfigure}\hfill\begin{subfigure}{0.095\linewidth}
  \begin{center}
    %\fbox{\rule{0pt}{2in} \rule{\linewidth}{0pt}}
    \includegraphics[width=\linewidth]{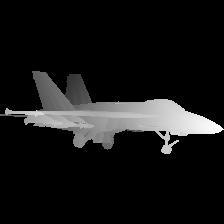}
    %\caption{}
  \end{center}
  \end{subfigure}\hfill\begin{subfigure}{0.095\linewidth}
  \begin{center}
    %\fbox{\rule{0pt}{2in} \rule{\linewidth}{0pt}}
    \includegraphics[width=\linewidth]{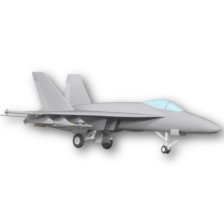}
    %\caption{}
  \end{center}
  \end{subfigure}\hfill\begin{subfigure}{0.095\linewidth}
  \begin{center}
    %\fbox{\rule{0pt}{2in} \rule{\linewidth}{0pt}}
    \includegraphics[width=\linewidth]{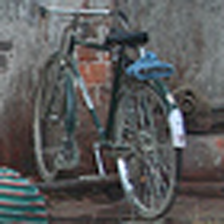}
    %\caption{}
  \end{center}
  \end{subfigure}\hfill\begin{subfigure}{0.095\linewidth}
  \begin{center}
    %\fbox{\rule{0pt}{2in} \rule{\linewidth}{0pt}}
    \includegraphics[width=\linewidth]{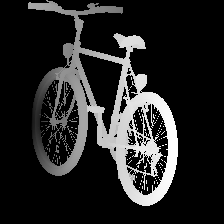}
    %\caption{}
  \end{center}
  \end{subfigure}\hfill\begin{subfigure}{0.095\linewidth}
  \begin{center}
    %\fbox{\rule{0pt}{2in} \rule{\linewidth}{0pt}}
    \includegraphics[width=\linewidth]{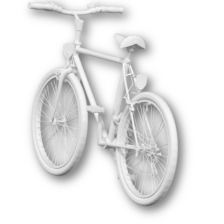}
    %\caption{}
  \end{center}
  \end{subfigure}\hfill\begin{subfigure}{0.095\linewidth}
  \begin{center}
    %\fbox{\rule{0pt}{2in} \rule{\linewidth}{0pt}}
    \includegraphics[width=\linewidth]{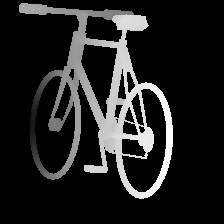}
    %\caption{}
  \end{center}
  \end{subfigure}\hfill\begin{subfigure}{0.095\linewidth}
  \begin{center}
    %\fbox{\rule{0pt}{2in} \rule{\linewidth}{0pt}}
    \includegraphics[width=\linewidth]{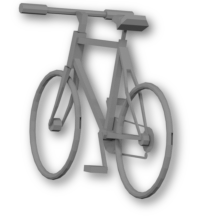}
    %\caption{}
  \end{center}
  \end{subfigure}\\[0.06cm]
  \begin{subfigure}{0.095\linewidth}
    \begin{center}
      %\fbox{\rule{0pt}{2in} \rule{\linewidth}{0pt}}
      \includegraphics[width=\linewidth]{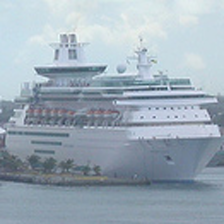}
      %\caption{}
    \end{center}
  \end{subfigure}\hfill\begin{subfigure}{0.095\linewidth}
  \begin{center}
    %\fbox{\rule{0pt}{2in} \rule{\linewidth}{0pt}}
    \includegraphics[width=\linewidth]{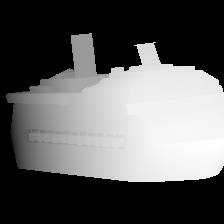}
    %\caption{}
  \end{center}
  \end{subfigure}\hfill\begin{subfigure}{0.095\linewidth}
  \begin{center}
    %\fbox{\rule{0pt}{2in} \rule{\linewidth}{0pt}}
    \includegraphics[width=\linewidth]{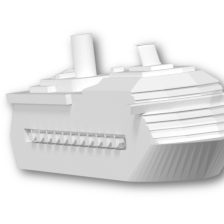}
    %\caption{}
  \end{center}
  \end{subfigure}\hfill\begin{subfigure}{0.095\linewidth}
  \begin{center}
    %\fbox{\rule{0pt}{2in} \rule{\linewidth}{0pt}}
    \includegraphics[width=\linewidth]{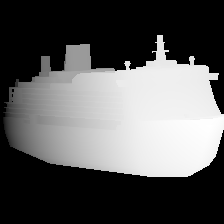}
    %\caption{}
  \end{center}
  \end{subfigure}\hfill\begin{subfigure}{0.095\linewidth}
  \begin{center}
    %\fbox{\rule{0pt}{2in} \rule{\linewidth}{0pt}}
    \includegraphics[width=\linewidth]{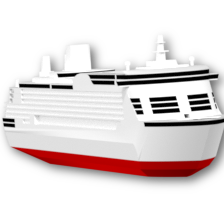}
    %\caption{}
  \end{center}
  \end{subfigure}\hfill\begin{subfigure}{0.095\linewidth}
  \begin{center}
    %\fbox{\rule{0pt}{2in} \rule{\linewidth}{0pt}}
    \includegraphics[width=\linewidth]{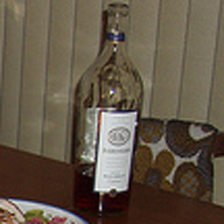}
    %\caption{}
  \end{center}
  \end{subfigure}\hfill\begin{subfigure}{0.095\linewidth}
  \begin{center}
    %\fbox{\rule{0pt}{2in} \rule{\linewidth}{0pt}}
    \includegraphics[width=\linewidth]{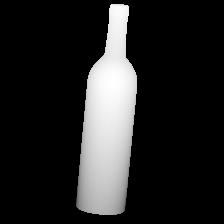}
    %\caption{}
  \end{center}
  \end{subfigure}\hfill\begin{subfigure}{0.095\linewidth}
  \begin{center}
    %\fbox{\rule{0pt}{2in} \rule{\linewidth}{0pt}}
    \includegraphics[width=\linewidth]{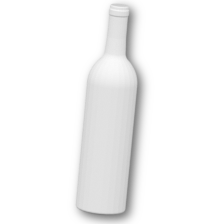}
    %\caption{}
  \end{center}
  \end{subfigure}\hfill\begin{subfigure}{0.095\linewidth}
  \begin{center}
    %\fbox{\rule{0pt}{2in} \rule{\linewidth}{0pt}}
    \includegraphics[width=\linewidth]{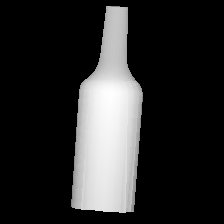}
    %\caption{}
  \end{center}
  \end{subfigure}\hfill\begin{subfigure}{0.095\linewidth}
  \begin{center}
    %\fbox{\rule{0pt}{2in} \rule{\linewidth}{0pt}}
    \includegraphics[width=\linewidth]{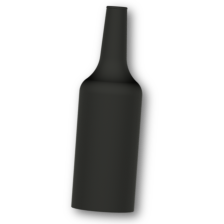}
    %\caption{}
  \end{center}
  \end{subfigure}\\[0.06cm]
  \begin{subfigure}{0.095\linewidth}
    \begin{center}
      %\fbox{\rule{0pt}{2in} \rule{\linewidth}{0pt}}
      \includegraphics[width=\linewidth]{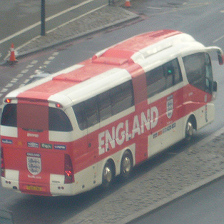}
      %\caption{}
    \end{center}
  \end{subfigure}\hfill\begin{subfigure}{0.095\linewidth}
  \begin{center}
    %\fbox{\rule{0pt}{2in} \rule{\linewidth}{0pt}}
    \includegraphics[width=\linewidth]{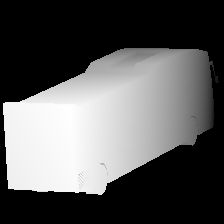}
    %\caption{}
  \end{center}
  \end{subfigure}\hfill\begin{subfigure}{0.095\linewidth}
  \begin{center}
    %\fbox{\rule{0pt}{2in} \rule{\linewidth}{0pt}}
    \includegraphics[width=\linewidth]{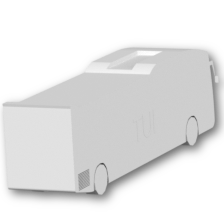}
    %\caption{}
  \end{center}
  \end{subfigure}\hfill\begin{subfigure}{0.095\linewidth}
  \begin{center}
    %\fbox{\rule{0pt}{2in} \rule{\linewidth}{0pt}}
    \includegraphics[width=\linewidth]{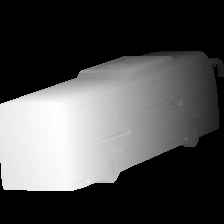}
    %\caption{}
  \end{center}
  \end{subfigure}\hfill\begin{subfigure}{0.095\linewidth}
  \begin{center}
    %\fbox{\rule{0pt}{2in} \rule{\linewidth}{0pt}}
    \includegraphics[width=\linewidth]{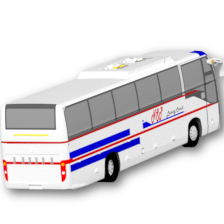}
    %\caption{}
  \end{center}
  \end{subfigure}\hfill\begin{subfigure}{0.095\linewidth}
  \begin{center}
    %\fbox{\rule{0pt}{2in} \rule{\linewidth}{0pt}}
    \includegraphics[width=\linewidth]{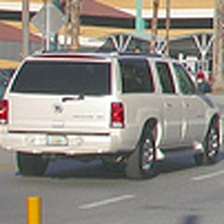}
    %\caption{}
  \end{center}
  \end{subfigure}\hfill\begin{subfigure}{0.095\linewidth}
  \begin{center}
    %\fbox{\rule{0pt}{2in} \rule{\linewidth}{0pt}}
    \includegraphics[width=\linewidth]{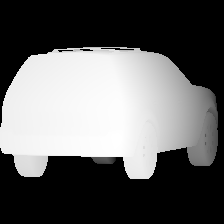}
    %\caption{}
  \end{center}
  \end{subfigure}\hfill\begin{subfigure}{0.095\linewidth}
  \begin{center}
    %\fbox{\rule{0pt}{2in} \rule{\linewidth}{0pt}}
    \includegraphics[width=\linewidth]{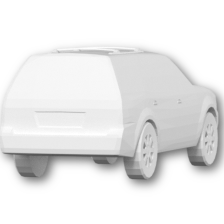}
    %\caption{}
  \end{center}
  \end{subfigure}\hfill\begin{subfigure}{0.095\linewidth}
  \begin{center}
    %\fbox{\rule{0pt}{2in} \rule{\linewidth}{0pt}}
    \includegraphics[width=\linewidth]{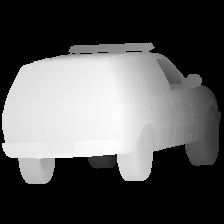}
    %\caption{}
  \end{center}
  \end{subfigure}\hfill\begin{subfigure}{0.095\linewidth}
  \begin{center}
    %\fbox{\rule{0pt}{2in} \rule{\linewidth}{0pt}}
    \includegraphics[width=\linewidth]{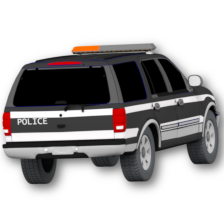}
    %\caption{}
  \end{center}
  \end{subfigure}\\[0.06cm]
  \begin{subfigure}{0.095\linewidth}
    \begin{center}
      %\fbox{\rule{0pt}{2in} \rule{\linewidth}{0pt}}
      \includegraphics[width=\linewidth]{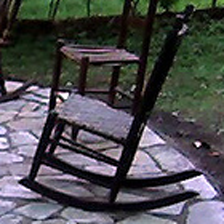}
      %\caption{}
    \end{center}
  \end{subfigure}\hfill\begin{subfigure}{0.095\linewidth}
  \begin{center}
    %\fbox{\rule{0pt}{2in} \rule{\linewidth}{0pt}}
    \includegraphics[width=\linewidth]{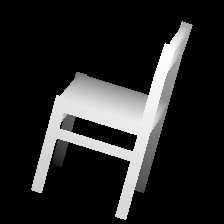}
    %\caption{}
  \end{center}
  \end{subfigure}\hfill\begin{subfigure}{0.095\linewidth}
  \begin{center}
    %\fbox{\rule{0pt}{2in} \rule{\linewidth}{0pt}}
    \includegraphics[width=\linewidth]{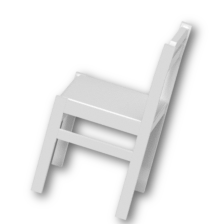}
    %\caption{}
  \end{center}
  \end{subfigure}\hfill\begin{subfigure}{0.095\linewidth}
  \begin{center}
    %\fbox{\rule{0pt}{2in} \rule{\linewidth}{0pt}}
    \includegraphics[width=\linewidth]{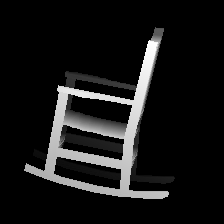}
    %\caption{}
  \end{center}
  \end{subfigure}\hfill\begin{subfigure}{0.095\linewidth}
  \begin{center}
    %\fbox{\rule{0pt}{2in} \rule{\linewidth}{0pt}}
    \includegraphics[width=\linewidth]{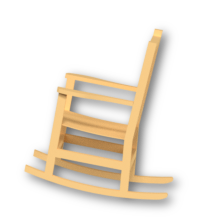}
    %\caption{}
  \end{center}
  \end{subfigure}\hfill\begin{subfigure}{0.095\linewidth}
  \begin{center}
    %\fbox{\rule{0pt}{2in} \rule{\linewidth}{0pt}}
    \includegraphics[width=\linewidth]{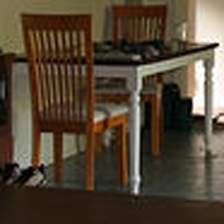}
    %\caption{}
  \end{center}
  \end{subfigure}\hfill\begin{subfigure}{0.095\linewidth}
  \begin{center}
    %\fbox{\rule{0pt}{2in} \rule{\linewidth}{0pt}}
    \includegraphics[width=\linewidth]{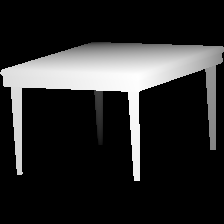}
    %\caption{}
  \end{center}
  \end{subfigure}\hfill\begin{subfigure}{0.095\linewidth}
  \begin{center}
    %\fbox{\rule{0pt}{2in} \rule{\linewidth}{0pt}}
    \includegraphics[width=\linewidth]{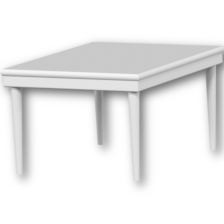}
    %\caption{}
  \end{center}
  \end{subfigure}\hfill\begin{subfigure}{0.095\linewidth}
  \begin{center}
    %\fbox{\rule{0pt}{2in} \rule{\linewidth}{0pt}}
    \includegraphics[width=\linewidth]{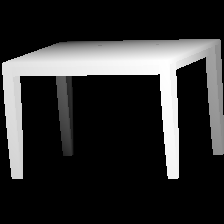}
    %\caption{}
  \end{center}
  \end{subfigure}\hfill\begin{subfigure}{0.095\linewidth}
  \begin{center}
    %\fbox{\rule{0pt}{2in} \rule{\linewidth}{0pt}}
    \includegraphics[width=\linewidth]{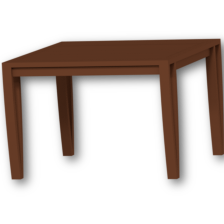}
    %\caption{}
  \end{center}
  \end{subfigure}\\[0.06cm]
  \begin{subfigure}{0.095\linewidth}
    \begin{center}
      %\fbox{\rule{0pt}{2in} \rule{\linewidth}{0pt}}
      \includegraphics[width=\linewidth]{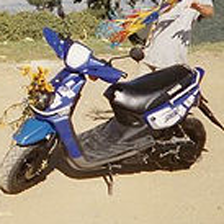}
      %\caption{}
    \end{center}
  \end{subfigure}\hfill\begin{subfigure}{0.095\linewidth}
  \begin{center}
    %\fbox{\rule{0pt}{2in} \rule{\linewidth}{0pt}}
    \includegraphics[width=\linewidth]{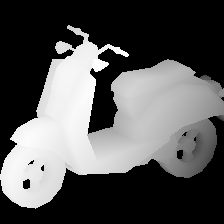}
    %\caption{}
  \end{center}
  \end{subfigure}\hfill\begin{subfigure}{0.095\linewidth}
  \begin{center}
    %\fbox{\rule{0pt}{2in} \rule{\linewidth}{0pt}}
    \includegraphics[width=\linewidth]{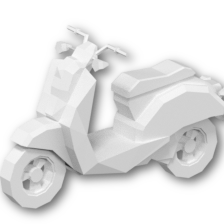}
    %\caption{}
  \end{center}
  \end{subfigure}\hfill\begin{subfigure}{0.095\linewidth}
  \begin{center}
    %\fbox{\rule{0pt}{2in} \rule{\linewidth}{0pt}}
    \includegraphics[width=\linewidth]{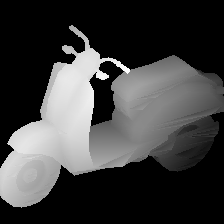}
    %\caption{}
  \end{center}
  \end{subfigure}\hfill\begin{subfigure}{0.095\linewidth}
  \begin{center}
    %\fbox{\rule{0pt}{2in} \rule{\linewidth}{0pt}}
    \includegraphics[width=\linewidth]{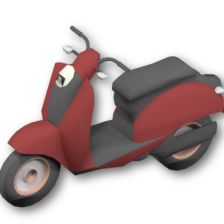}
    %\caption{}
  \end{center}
  \end{subfigure}\hfill\begin{subfigure}{0.095\linewidth}
  \begin{center}
    %\fbox{\rule{0pt}{2in} \rule{\linewidth}{0pt}}
    \includegraphics[width=\linewidth]{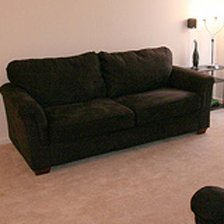}
    %\caption{}
  \end{center}
  \end{subfigure}\hfill\begin{subfigure}{0.095\linewidth}
  \begin{center}
    %\fbox{\rule{0pt}{2in} \rule{\linewidth}{0pt}}
    \includegraphics[width=\linewidth]{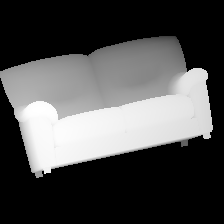}
    %\caption{}
  \end{center}
  \end{subfigure}\hfill\begin{subfigure}{0.095\linewidth}
  \begin{center}
    %\fbox{\rule{0pt}{2in} \rule{\linewidth}{0pt}}
    \includegraphics[width=\linewidth]{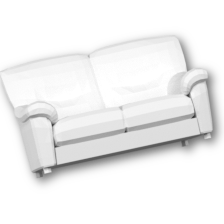}
    %\caption{}
  \end{center}
  \end{subfigure}\hfill\begin{subfigure}{0.095\linewidth}
  \begin{center}
    %\fbox{\rule{0pt}{2in} \rule{\linewidth}{0pt}}
    \includegraphics[width=\linewidth]{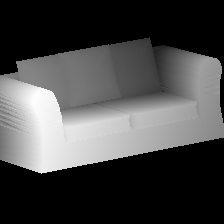}
    %\caption{}
  \end{center}
  \end{subfigure}\hfill\begin{subfigure}{0.095\linewidth}
  \begin{center}
    %\fbox{\rule{0pt}{2in} \rule{\linewidth}{0pt}}
    \includegraphics[width=\linewidth]{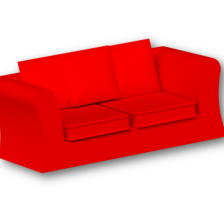}
    %\caption{}
  \end{center}
  \end{subfigure}\\[0.06cm]
  \begin{subfigure}{0.095\linewidth}
    \begin{center}
      %\fbox{\rule{0pt}{2in} \rule{\linewidth}{0pt}}
      \includegraphics[width=\linewidth]{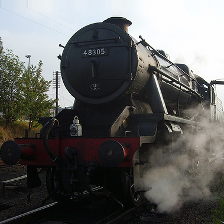}
      %\caption{}
    \end{center}
  \end{subfigure}\hfill\begin{subfigure}{0.095\linewidth}
  \begin{center}
    %\fbox{\rule{0pt}{2in} \rule{\linewidth}{0pt}}
    \includegraphics[width=\linewidth]{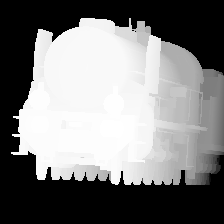}
    %\caption{}
  \end{center}
  \end{subfigure}\hfill\begin{subfigure}{0.095\linewidth}
  \begin{center}
    %\fbox{\rule{0pt}{2in} \rule{\linewidth}{0pt}}
    \includegraphics[width=\linewidth]{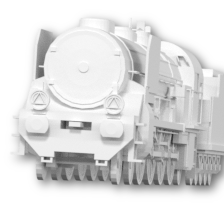}
    %\caption{}
  \end{center}
  \end{subfigure}\hfill\begin{subfigure}{0.095\linewidth}
  \begin{center}
    %\fbox{\rule{0pt}{2in} \rule{\linewidth}{0pt}}
    \includegraphics[width=\linewidth]{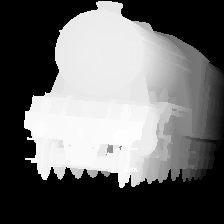}
    %\caption{}
  \end{center}
  \end{subfigure}\hfill\begin{subfigure}{0.095\linewidth}
  \begin{center}
    %\fbox{\rule{0pt}{2in} \rule{\linewidth}{0pt}}
    \includegraphics[width=\linewidth]{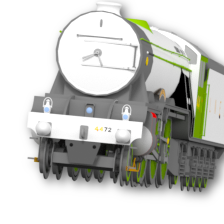}
    %\caption{}
  \end{center}
  \end{subfigure}\hfill\begin{subfigure}{0.095\linewidth}
  \begin{center}
    %\fbox{\rule{0pt}{2in} \rule{\linewidth}{0pt}}
    \includegraphics[width=\linewidth]{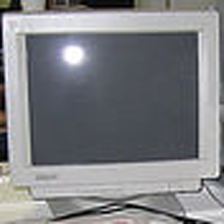}
    %\caption{}
  \end{center}
  \end{subfigure}\hfill\begin{subfigure}{0.095\linewidth}
  \begin{center}
    %\fbox{\rule{0pt}{2in} \rule{\linewidth}{0pt}}
    \includegraphics[width=\linewidth]{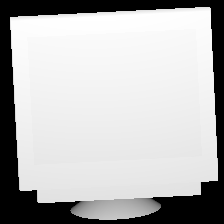}
    %\caption{}
  \end{center}
  \end{subfigure}\hfill\begin{subfigure}{0.095\linewidth}
  \begin{center}
    %\fbox{\rule{0pt}{2in} \rule{\linewidth}{0pt}}
    \includegraphics[width=\linewidth]{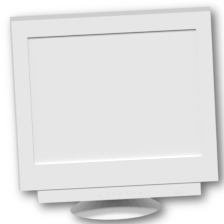}
    %\caption{}
  \end{center}
  \end{subfigure}\hfill\begin{subfigure}{0.095\linewidth}
  \begin{center}
    %\fbox{\rule{0pt}{2in} \rule{\linewidth}{0pt}}
    \includegraphics[width=\linewidth]{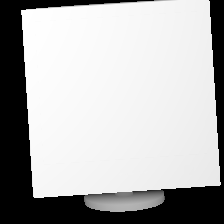}
    %\caption{}
  \end{center}
  \end{subfigure}\hfill\begin{subfigure}{0.095\linewidth}
  \begin{center}
    %\fbox{\rule{0pt}{2in} \rule{\linewidth}{0pt}}
    \includegraphics[width=\linewidth]{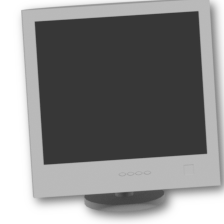}
    %\caption{}
  \end{center}
  \end{subfigure}
  \caption{Qualitative results for 3D pose estimation and 3D model retrieval from ShapeNet given images from Pascal3D+ for all twelve categories. For each category, we show: the query RGB image; the depth image and RGB rendering of the ground truth 3D model from Pascal3D+ under the ground truth pose from Pascal3D+; the depth image and RGB rendering of our retrieved 3D model from ShapeNet under our predicted pose. We provide more results in the supplementary material.}
  \label{fig:retrieval}
  \vspace{-0.5cm}
\end{figure*}

The low $Acc_{\frac{\pi}{6}}$ scores for the category \textit{table} can be explained by three factors. First, many tables are partly occluded by chairs (see \textit{table} in Fig.~\ref{fig:retrieval}). Second, the evaluation protocol does not take into account that many tables are ambiguous with respect to an azimuth rotation of $\pi$, $\frac{\pi}{2}$ or even have an axis of symmetry, \eg, a round table. In some cases, our system predicts an ambiguous pose instead of the ground truth pose, while it is not possible to differentiate between the two poses. The evaluation protocol needs to be changed to take this into account. Last, the number of validation samples is very small (\ie, $21$) and, therefore, the reported results for this category are highly biased.

\subsubsection{Category-Agnostic Pose Estimation}

So far, the discussed results are category-specific, which means that the ground truth category must be known at runtime. In fact, all methods use a separate output layer for each category. However, our approach is able to make category-agnostic predictions which generalize across different categories. In this case, we use a single 19 neuron output layer which is shared for all categories making our approach scalable. Our category-agnostic pose estimation even outperforms the previous category-specific state-of-the-art for some categories, because it fully leverages the mutual information between similar categories like \textit{bike} and \textit{mbike}, for example, as shown in Table~\ref{table:viewpoint-sota}.

\definecolor{lightgreen}{RGB}{245,255,245}
\begin{table*}[!t]
	\centering
	\begin{tabular}{lrrrrrrrrrrrr|X>{\columncolor{lightgreen}}c}
		\toprule
		&\multicolumn{1}{c}{aero}&\multicolumn{1}{c}{bike}&\multicolumn{1}{c}{boat}&\multicolumn{1}{c}{bottle}&\multicolumn{1}{c}{bus}&\multicolumn{1}{c}{car}&\multicolumn{1}{c}{chair}&\multicolumn{1}{c}{table}&\multicolumn{1}{c}{mbike}&\multicolumn{1}{c}{sofa}&\multicolumn{1}{c}{train}&\multicolumn{1}{c|}{tv}&\multicolumn{1}{c}{\cellcolor{lightgreen}mean}\\
		\midrule
		\textit{Top-1-Acc} (Rand)&0.15&0.21&0.36&0.25&0.25&0.10&0.15&0.10&0.28&0.31&0.27&0.27&0.2250\\
		\textit{Top-1-Acc} (Cano)&0.12&0.25&0.38&0.35&0.45&0.21&0.20&0.15&0.20&0.21&0.49&0.50&0.2925\\
		\textit{Top-1-Acc} (Off)&0.48&0.33&0.58&\bf0.41&0.75&0.35&0.28&0.10&0.44&0.28&0.62&0.63&0.4375\\
		\textit{Top-1-Acc} (Pred)&0.48&0.31&\bf0.60&\bf0.41&0.78&0.41&0.29&0.19&0.43&\bf0.36&0.65&0.61&0.4600\\
		\textit{Top-1-Acc} (GT)&\bf0.53&\bf0.38&0.51&0.37&\bf0.79&\bf0.44&\bf0.32&\bf0.43&\bf0.48&0.33&\bf0.66&\bf0.72&\bf0.4967\\
		\bottomrule
	\end{tabular}
	\caption{3D model retrieval accuracy using ground truth detections on Pascal3D+.}
	\label{table:retrieval}
	\vspace{-4pt}
\end{table*}

\subsection{3D Model Retrieval}
\label{sec:exp-retrieval}

Now we demonstrate our 3D model retrieval approach using our predicted pose. First, we present a quantitative evaluation of our approach on Pascal3D+. Second, we show qualitative results for 3D model retrieval from ShapeNet given images from Pascal3D+. Finally, we use our predicted 6-DoF pose and 3-DoF dimensions to precisely align retrieved 3D models with objects in real world images.

\subsubsection{3D Model Retrieval from Pascal3D+}

Since Pascal3D+ provides correspondences between RGB images and 3D models as well as pose annotations, we can train our approach purely on this dataset. In fact, we are the first to report quantitative results for 3D model retrieval on this dataset. For this purpose, we compute the top-1-accuracy (\textit{Top-1-Acc}), \ie, the percentage of evaluated samples for which the top retrieved model equals the ground truth model. This task is not trivial, because many models in Pascal3D+ have similar geometry and are hard to distinguish. Thus, we evaluate our approach using five different pose setups, \ie, the ground truth pose (\emph{GT}), our predicted pose (\emph{Pred}), our predicted pose with offline pre-computed descriptors (\emph{Off}), a canonical pose (\emph{Cano}) and a random pose (\emph{Rand}). Table~\ref{table:retrieval} shows quantitative retrieval results.

As expected, we achieve the highest accuracy assuming the ground truth pose to be known (\emph{GT}). In this case, our approach chooses the same 3D models as human annotators for 50\% of the validation images on average. However, if we render the 3D models under our predicted pose (\emph{Pred}), we almost match the accuracy of the ground truth pose setup. For some categories, we observe even better accuracy when using our predicted pose. This proves the high quality of our predicted pose. Moreover, our approach is fast and scalable at runtime while almost maintaining accuracy by using offline pre-computed descriptors (\emph{Off}). For this experiment, we discretize the pose space in intervals of $10^\circ$ and pre-compute descriptors for the 3D models. At runtime, we only match pre-computed descriptors from the discretized pose which is closest to our predicted pose and do not have to render 3D models online.
If we, in contrast, just render the 3D models under a random pose (\emph{Rand}) the performance decreases significantly. Rendering models under a frontal view (\emph{Cano}) on the other hand provides a useful bias for the categories \textit{train}, \textit{bus} and \textit{tv monitor} which are frequently seen from an almost frontal view in this dataset. These results confirm the importance of fine pose estimation in our approach.

\begin{figure}
	\begin{subfigure}{0.19\linewidth}
		\begin{center}
			%\fbox{\rule{0pt}{2in} \rule{\linewidth}{0pt}}
			\includegraphics[width=\linewidth]{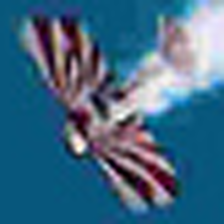}
			%\caption{}
		\end{center}
	\end{subfigure}\hfill\begin{subfigure}{0.19\linewidth}
		\begin{center}
			%\fbox{\rule{0pt}{2in} \rule{\linewidth}{0pt}}
			\includegraphics[width=\linewidth]{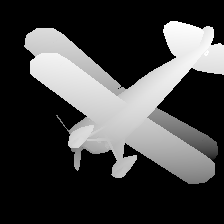}
			%\caption{}
		\end{center}
	\end{subfigure}\hfill\begin{subfigure}{0.19\linewidth}
		\begin{center}
			%\fbox{\rule{0pt}{2in} \rule{\linewidth}{0pt}}
			\includegraphics[width=\linewidth]{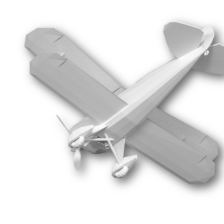}
			%\caption{}
		\end{center}
	\end{subfigure}\hfill\begin{subfigure}{0.19\linewidth}
		\begin{center}
			%\fbox{\rule{0pt}{2in} \rule{\linewidth}{0pt}}
			\includegraphics[width=\linewidth]{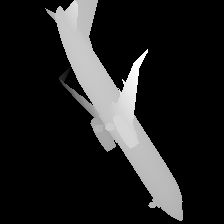}
			%\caption{}
		\end{center}
	\end{subfigure}\hfill\begin{subfigure}{0.19\linewidth}
		\begin{center}
			%\fbox{\rule{0pt}{2in} \rule{\linewidth}{0pt}}
			\includegraphics[width=\linewidth]{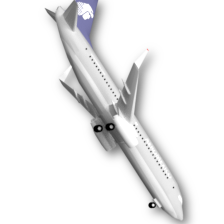}
			%\caption{}
		\end{center}
	\end{subfigure}\\[0.07cm]
\begin{subfigure}{0.19\linewidth}
		\begin{center}
			%\fbox{\rule{0pt}{2in} \rule{\linewidth}{0pt}}
			\includegraphics[width=\linewidth]{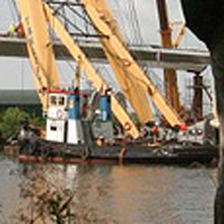}
			%\caption{}
		\end{center}
	\end{subfigure}\hfill\begin{subfigure}{0.19\linewidth}
		\begin{center}
			%\fbox{\rule{0pt}{2in} \rule{\linewidth}{0pt}}
			\includegraphics[width=\linewidth]{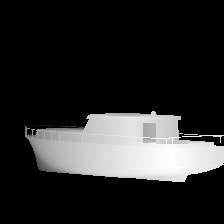}
			%\caption{}
		\end{center}
	\end{subfigure}\hfill\begin{subfigure}{0.19\linewidth}
		\begin{center}
			%\fbox{\rule{0pt}{2in} \rule{\linewidth}{0pt}}
			\includegraphics[width=\linewidth]{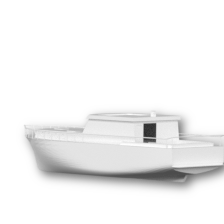}
			%\caption{}
		\end{center}
	\end{subfigure}\hfill\begin{subfigure}{0.19\linewidth}
		\begin{center}
			%\fbox{\rule{0pt}{2in} \rule{\linewidth}{0pt}}
			\includegraphics[width=\linewidth]{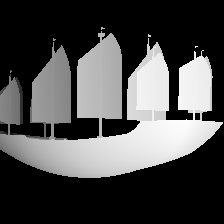}
			%\caption{}
		\end{center}
	\end{subfigure}\hfill\begin{subfigure}{0.19\linewidth}
		\begin{center}
			%\fbox{\rule{0pt}{2in} \rule{\linewidth}{0pt}}
			\includegraphics[width=\linewidth]{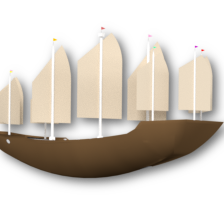}
			%\caption{}
		\end{center}
	\end{subfigure}
	\caption{Example failure cases of our 3D model retrieval approach (same image arrangement as in Fig.~\ref{fig:retrieval}). Top: The pose estimation fails because no similar pose was seen during training, as a result, the model retrieval fails. In this case, also the ground truth pose annotation from Pascal3D+ is not accurate. Bottom: While we estimate the pose correctly, the model retrieval fails due to heavy clutter.}
	\label{fig:failure}
	\vspace{-0.1cm}
\end{figure}

\begin{figure}
\begin{subfigure}{0.32\linewidth} \begin{center}
            %\fbox{\rule{0pt}{2in} \rule{\linewidth}{0pt}}
            \includegraphics[width=\linewidth]{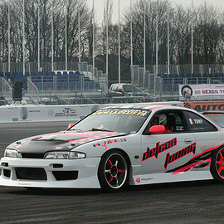}
            %\caption{}
        \end{center} \end{subfigure}\hfill\begin{subfigure}{0.32\linewidth}
        \begin{center}
        %\fbox{\rule{0pt}{2in} \rule{\linewidth}{0pt}}
        \includegraphics[width=\linewidth]{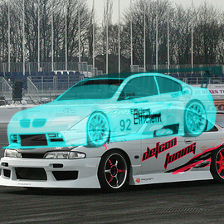}
        %\caption{}
    \end{center} \end{subfigure}\hfill\begin{subfigure}{0.32\linewidth}
    \begin{center}
            %\fbox{\rule{0pt}{2in} \rule{\linewidth}{0pt}}
            \includegraphics[width=\linewidth]{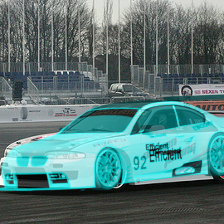}
            %\caption{}
        \end{center} \end{subfigure}\\[0.13cm] \begin{subfigure}{0.32\linewidth}
        \begin{center}
            %\fbox{\rule{0pt}{2in} \rule{\linewidth}{0pt}}
            \includegraphics[width=\linewidth]{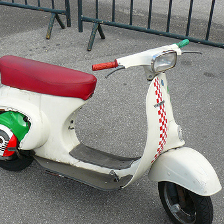}
            %\caption{}
        \end{center} \end{subfigure}\hfill\begin{subfigure}{0.32\linewidth}
        \begin{center}
            %\fbox{\rule{0pt}{2in} \rule{\linewidth}{0pt}}
            \includegraphics[width=\linewidth]{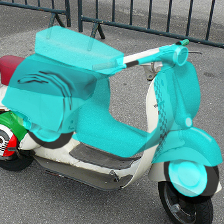}
            %\caption{}
        \end{center} \end{subfigure}\hfill\begin{subfigure}{0.32\linewidth}
        \begin{center}
            %\fbox{\rule{0pt}{2in} \rule{\linewidth}{0pt}}
            \includegraphics[width=\linewidth]{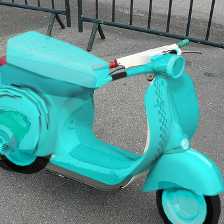}
            %\caption{}
        \end{center} \end{subfigure} \caption{We use our predicted 6-DoF pose
        and 3-DoF dimensions to refine the alignment between the object and a
        rendering. Left: A detected object, which is not centered on the image
        window. Middle: A rendering which just uses our predicted 3-DoF rotation. Right:
        A rendering which uses our predicted 6-DoF pose and 3-DoF dimensions.}
        \label{fig:alignment}
        \vspace{-0.35cm}
\end{figure}

\vspace{-0.1cm}
\subsubsection{3D Model Retrieval from ShapeNet}

In contrast to Pascal3D+, ShapeNet provides a significantly larger spectrum of 3D models. Thus, we now evaluate our retrieval approach trained purely on Pascal3D+ for 3D model retrieval from ShapeNet given previously unseen images from Pascal3D+. Fig.~\ref{fig:retrieval} shows qualitative retrieval results for all twelve categories. Our approach predicts accurate 3D poses and 3D models for objects of different categories. In some cases, our predicted pose (see \textit{sofa} in Fig.~\ref{fig:retrieval}) or our retrieved model from ShapeNet (see \textit{aeroplane} and \textit{chair} in Fig.~\ref{fig:retrieval}) are even more accurate than the annotated ground truth from Pascal3D+. While the geometry of the retrieved models corresponds well to the objects in the query images, the materials and textures typically do not. The reason for this is that we use depth images for retrieval, which do not include color information. This issue can be addressed by extracting texture information from the query RGB image or by performing retrieval with RGBD images. However, this is up to future research. Fig.~\ref{fig:failure} shows failure cases of our approach. If the pose estimation fails, the model retrieval becomes even more difficult. This is also reflected in Table~\ref{table:retrieval}, where we observe a strong decrease in performance when we render models without pose information (\emph{Rand} and \emph{Cano}). Also, if there is too much clutter in the query image, we cannot retrieve an accurate 3D model.

\vspace{-0.1cm}

\subsubsection{3D Model Alignment}

Finally, we use our predicted 6-DoF pose and 3-DoF dimensions to precisely align retrieved 3D models with objects in real world images. Fig.~\ref{fig:alignment} shows how we improve the 2D object localization and the alignment between the object and a rendering using our predicted pose and dimensions. This is especially useful if the object detection is not fully accurate, which is true in almost all situations. In this case, the detected image windows are a bit too small and the objects are not centered in the image windows. Thus, if we just render a model under our predicted rotation, re-scale it to tightly fit into the 2D image window, and center it in the 2D image window, the alignment is poor. However, if we additionally use our predicted translation and 3D dimensions for scaling and positioning, we significantly improve the alignment between the object and the rendering. This is of tremendous importance for robotics or augmented reality applications.

%\vspace{-0.25cm}
\section{Conclusion}

3D object retrieval from RGB images in the wild is an important but challenging task. Existing approaches address this problem by training on vast amounts of synthetic data. However, there is a significant domain gap between real and synthetic images which limits performance. For this reason, we learn to map real RGB images and synthetic depth images to a common representation. Additionally, we show that estimating the object pose is a useful prior for 3D model retrieval. Our approach is scalable as it supports category-agnostic predictions and offline pre-computed descriptors. We do not only outperform the state-of-the-art for viewpoint estimation on Pascal3D+, but also retrieve accurate 3D models from ShapeNet given unseen RGB images from Pascal3D+. Finally, these results motivate future research on jointly learning from real and synthetic data.

\vspace{-0.25cm}
\paragraph{Acknowledgement} This work was funded by the Christian Doppler Laboratory for Semantic 3D Computer Vision.

{\small
	\bibliographystyle{ieee}
	\bibliography{string,cleaned_references}
}

\cleardoublepage

\twocolumn[{
	\newpage
	\null
	\vskip .375in
	\begin{center}
		{\Large \bf 3D Pose Estimation and 3D Model Retrieval for Objects in the Wild\newline Supplementary Material \par}
		% additional two empty lines at the end of the title
		\vspace*{24pt}
		{
			\large
			\lineskip .5em
			\begin{tabular}[t]{c}
				
			\end{tabular}
			\par
		}
		% additional small space at the end of the author name
		\vskip .5em
		% additional empty line at the end of the title block
		\vspace*{-11pt}
	\end{center}
}]

%% Supplementary

In the following, we provide additional qualitative results for our 3D model retrieval approach in Sec.~\ref{sec:retrieval}, which complement those presented in the paper. Furthermore, we analyze failure cases for both 3D model retrieval and the underlying 3D pose estimation in Sec.~\ref{sec:fc}. Finally, in Sec.~\ref{sec:details} we discuss implementation details, parameter choices, and other relevant settings.

\section{3D Model Retrieval}
\label{sec:retrieval}

\begin{figure*}
  \begin{subfigure}{0.095\linewidth}
    \begin{center}
      %\fbox{\rule{0pt}{2in} \rule{\linewidth}{0pt}}
      \includegraphics[width=\linewidth]{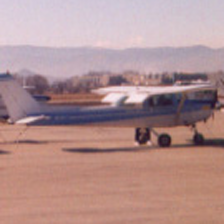}
      %\caption{}
    \end{center}
  \end{subfigure}\hfill\begin{subfigure}{0.095\linewidth}
  \begin{center}
    %\fbox{\rule{0pt}{2in} \rule{\linewidth}{0pt}}
    \includegraphics[width=\linewidth]{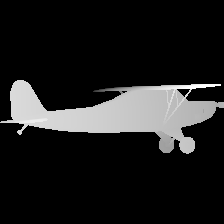}
    %\caption{}
  \end{center}
  \end{subfigure}\hfill\begin{subfigure}{0.095\linewidth}
  \begin{center}
    %\fbox{\rule{0pt}{2in} \rule{\linewidth}{0pt}}
    \includegraphics[width=\linewidth]{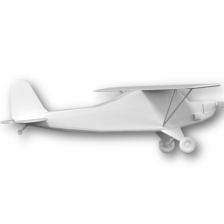}
    %\caption{}
  \end{center}
  \end{subfigure}\hfill\begin{subfigure}{0.095\linewidth}
  \begin{center}
    %\fbox{\rule{0pt}{2in} \rule{\linewidth}{0pt}}
    \includegraphics[width=\linewidth]{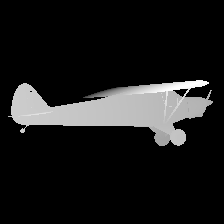}
    %\caption{}
  \end{center}
  \end{subfigure}\hfill\begin{subfigure}{0.095\linewidth}
  \begin{center}
    %\fbox{\rule{0pt}{2in} \rule{\linewidth}{0pt}}
    \includegraphics[width=\linewidth]{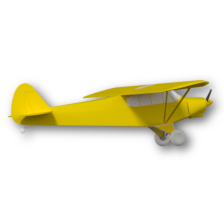}
    %\caption{}
  \end{center}
  \end{subfigure}\hfill\begin{subfigure}{0.095\linewidth}
  \begin{center}
    %\fbox{\rule{0pt}{2in} \rule{\linewidth}{0pt}}
    \includegraphics[width=\linewidth]{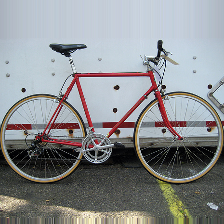}
    %\caption{}
  \end{center}
  \end{subfigure}\hfill\begin{subfigure}{0.095\linewidth}
  \begin{center}
    %\fbox{\rule{0pt}{2in} \rule{\linewidth}{0pt}}
    \includegraphics[width=\linewidth]{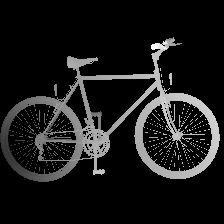}
    %\caption{}
  \end{center}
  \end{subfigure}\hfill\begin{subfigure}{0.095\linewidth}
  \begin{center}
    %\fbox{\rule{0pt}{2in} \rule{\linewidth}{0pt}}
    \includegraphics[width=\linewidth]{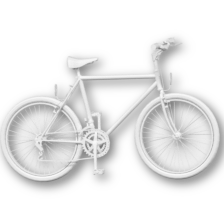}
    %\caption{}
  \end{center}
  \end{subfigure}\hfill\begin{subfigure}{0.095\linewidth}
  \begin{center}
    %\fbox{\rule{0pt}{2in} \rule{\linewidth}{0pt}}
    \includegraphics[width=\linewidth]{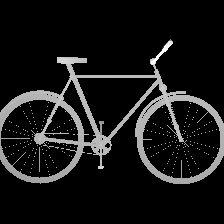}
    %\caption{}
  \end{center}
  \end{subfigure}\hfill\begin{subfigure}{0.095\linewidth}
  \begin{center}
    %\fbox{\rule{0pt}{2in} \rule{\linewidth}{0pt}}
    \includegraphics[width=\linewidth]{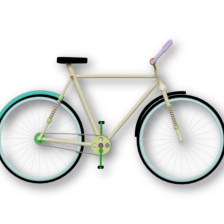}
    %\caption{}
  \end{center}
  \end{subfigure}\\[0.06cm]
  \begin{subfigure}{0.095\linewidth}
    \begin{center}
      %\fbox{\rule{0pt}{2in} \rule{\linewidth}{0pt}}
      \includegraphics[width=\linewidth]{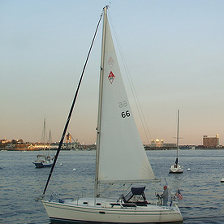}
      %\caption{}
    \end{center}
  \end{subfigure}\hfill\begin{subfigure}{0.095\linewidth}
  \begin{center}
    %\fbox{\rule{0pt}{2in} \rule{\linewidth}{0pt}}
    \includegraphics[width=\linewidth]{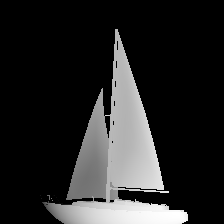}
    %\caption{}
  \end{center}
  \end{subfigure}\hfill\begin{subfigure}{0.095\linewidth}
  \begin{center}
    %\fbox{\rule{0pt}{2in} \rule{\linewidth}{0pt}}
    \includegraphics[width=\linewidth]{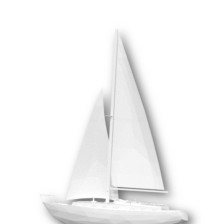}
    %\caption{}
  \end{center}
  \end{subfigure}\hfill\begin{subfigure}{0.095\linewidth}
  \begin{center}
    %\fbox{\rule{0pt}{2in} \rule{\linewidth}{0pt}}
    \includegraphics[width=\linewidth]{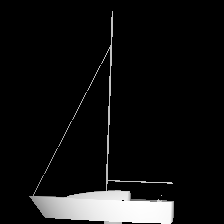}
    %\caption{}
  \end{center}
  \end{subfigure}\hfill\begin{subfigure}{0.095\linewidth}
  \begin{center}
    %\fbox{\rule{0pt}{2in} \rule{\linewidth}{0pt}}
    \includegraphics[width=\linewidth]{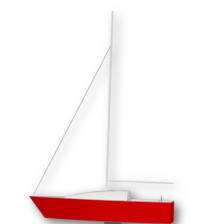}
    %\caption{}
  \end{center}
  \end{subfigure}\hfill\begin{subfigure}{0.095\linewidth}
  \begin{center}
    %\fbox{\rule{0pt}{2in} \rule{\linewidth}{0pt}}
    \includegraphics[width=\linewidth]{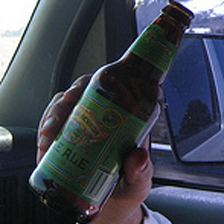}
    %\caption{}
  \end{center}
  \end{subfigure}\hfill\begin{subfigure}{0.095\linewidth}
  \begin{center}
    %\fbox{\rule{0pt}{2in} \rule{\linewidth}{0pt}}
    \includegraphics[width=\linewidth]{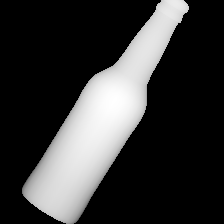}
    %\caption{}
  \end{center}
  \end{subfigure}\hfill\begin{subfigure}{0.095\linewidth}
  \begin{center}
    %\fbox{\rule{0pt}{2in} \rule{\linewidth}{0pt}}
    \includegraphics[width=\linewidth]{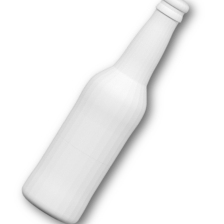}
    %\caption{}
  \end{center}
  \end{subfigure}\hfill\begin{subfigure}{0.095\linewidth}
  \begin{center}
    %\fbox{\rule{0pt}{2in} \rule{\linewidth}{0pt}}
    \includegraphics[width=\linewidth]{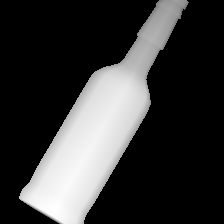}
    %\caption{}
  \end{center}
  \end{subfigure}\hfill\begin{subfigure}{0.095\linewidth}
  \begin{center}
    %\fbox{\rule{0pt}{2in} \rule{\linewidth}{0pt}}
    \includegraphics[width=\linewidth]{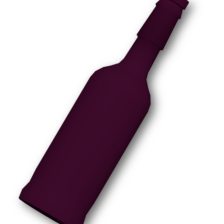}
    %\caption{}
  \end{center}
  \end{subfigure}\\[0.06cm]
  \begin{subfigure}{0.095\linewidth}
    \begin{center}
      %\fbox{\rule{0pt}{2in} \rule{\linewidth}{0pt}}
      \includegraphics[width=\linewidth]{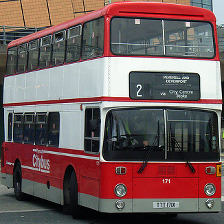}
      %\caption{}
    \end{center}
  \end{subfigure}\hfill\begin{subfigure}{0.095\linewidth}
  \begin{center}
    %\fbox{\rule{0pt}{2in} \rule{\linewidth}{0pt}}
    \includegraphics[width=\linewidth]{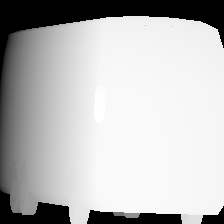}
    %\caption{}
  \end{center}
  \end{subfigure}\hfill\begin{subfigure}{0.095\linewidth}
  \begin{center}
    %\fbox{\rule{0pt}{2in} \rule{\linewidth}{0pt}}
    \includegraphics[width=\linewidth]{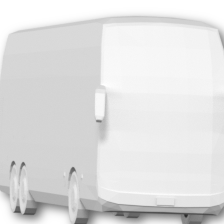}
    %\caption{}
  \end{center}
  \end{subfigure}\hfill\begin{subfigure}{0.095\linewidth}
  \begin{center}
    %\fbox{\rule{0pt}{2in} \rule{\linewidth}{0pt}}
    \includegraphics[width=\linewidth]{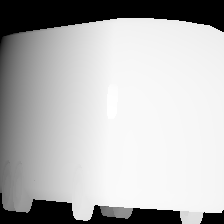}
    %\caption{}
  \end{center}
  \end{subfigure}\hfill\begin{subfigure}{0.095\linewidth}
  \begin{center}
    %\fbox{\rule{0pt}{2in} \rule{\linewidth}{0pt}}
    \includegraphics[width=\linewidth]{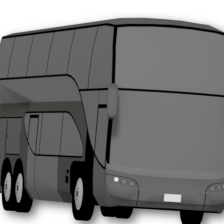}
    %\caption{}
  \end{center}
  \end{subfigure}\hfill\begin{subfigure}{0.095\linewidth}
  \begin{center}
    %\fbox{\rule{0pt}{2in} \rule{\linewidth}{0pt}}
    \includegraphics[width=\linewidth]{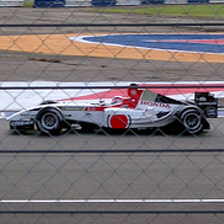}
    %\caption{}
  \end{center}
  \end{subfigure}\hfill\begin{subfigure}{0.095\linewidth}
  \begin{center}
    %\fbox{\rule{0pt}{2in} \rule{\linewidth}{0pt}}
    \includegraphics[width=\linewidth]{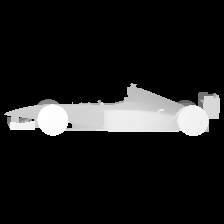}
    %\caption{}
  \end{center}
  \end{subfigure}\hfill\begin{subfigure}{0.095\linewidth}
  \begin{center}
    %\fbox{\rule{0pt}{2in} \rule{\linewidth}{0pt}}
    \includegraphics[width=\linewidth]{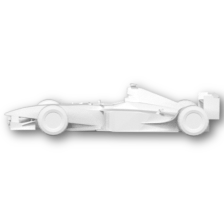}
    %\caption{}
  \end{center}
  \end{subfigure}\hfill\begin{subfigure}{0.095\linewidth}
  \begin{center}
    %\fbox{\rule{0pt}{2in} \rule{\linewidth}{0pt}}
    \includegraphics[width=\linewidth]{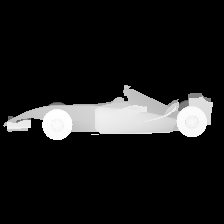}
    %\caption{}
  \end{center}
  \end{subfigure}\hfill\begin{subfigure}{0.095\linewidth}
  \begin{center}
    %\fbox{\rule{0pt}{2in} \rule{\linewidth}{0pt}}
    \includegraphics[width=\linewidth]{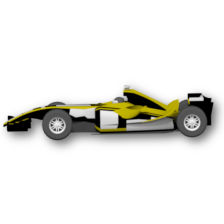}
    %\caption{}
  \end{center}
  \end{subfigure}\\[0.06cm]
  \begin{subfigure}{0.095\linewidth}
    \begin{center}
      %\fbox{\rule{0pt}{2in} \rule{\linewidth}{0pt}}
      \includegraphics[width=\linewidth]{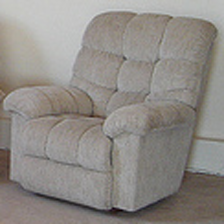}
      %\caption{}
    \end{center}
  \end{subfigure}\hfill\begin{subfigure}{0.095\linewidth}
  \begin{center}
    %\fbox{\rule{0pt}{2in} \rule{\linewidth}{0pt}}
    \includegraphics[width=\linewidth]{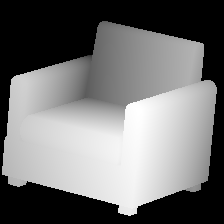}
    %\caption{}
  \end{center}
  \end{subfigure}\hfill\begin{subfigure}{0.095\linewidth}
  \begin{center}
    %\fbox{\rule{0pt}{2in} \rule{\linewidth}{0pt}}
    \includegraphics[width=\linewidth]{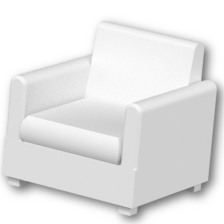}
    %\caption{}
  \end{center}
  \end{subfigure}\hfill\begin{subfigure}{0.095\linewidth}
  \begin{center}
    %\fbox{\rule{0pt}{2in} \rule{\linewidth}{0pt}}
    \includegraphics[width=\linewidth]{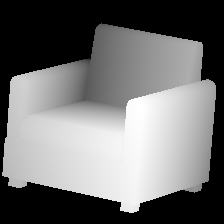}
    %\caption{}
  \end{center}
  \end{subfigure}\hfill\begin{subfigure}{0.095\linewidth}
  \begin{center}
    %\fbox{\rule{0pt}{2in} \rule{\linewidth}{0pt}}
    \includegraphics[width=\linewidth]{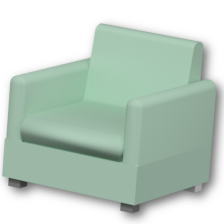}
    %\caption{}
  \end{center}
  \end{subfigure}\hfill\begin{subfigure}{0.095\linewidth}
  \begin{center}
    %\fbox{\rule{0pt}{2in} \rule{\linewidth}{0pt}}
    \includegraphics[width=\linewidth]{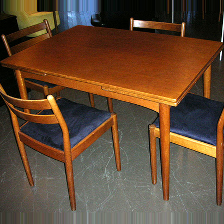}
    %\caption{}
  \end{center}
  \end{subfigure}\hfill\begin{subfigure}{0.095\linewidth}
  \begin{center}
    %\fbox{\rule{0pt}{2in} \rule{\linewidth}{0pt}}
    \includegraphics[width=\linewidth]{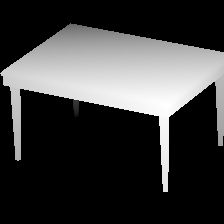}
    %\caption{}
  \end{center}
  \end{subfigure}\hfill\begin{subfigure}{0.095\linewidth}
  \begin{center}
    %\fbox{\rule{0pt}{2in} \rule{\linewidth}{0pt}}
    \includegraphics[width=\linewidth]{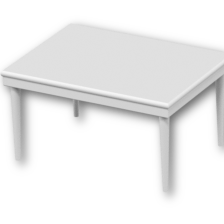}
    %\caption{}
  \end{center}
  \end{subfigure}\hfill\begin{subfigure}{0.095\linewidth}
  \begin{center}
    %\fbox{\rule{0pt}{2in} \rule{\linewidth}{0pt}}
    \includegraphics[width=\linewidth]{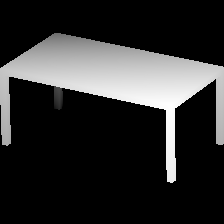}
    %\caption{}
  \end{center}
  \end{subfigure}\hfill\begin{subfigure}{0.095\linewidth}
  \begin{center}
    %\fbox{\rule{0pt}{2in} \rule{\linewidth}{0pt}}
    \includegraphics[width=\linewidth]{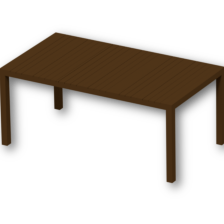}
    %\caption{}
  \end{center}
  \end{subfigure}\\[0.06cm]
  \begin{subfigure}{0.095\linewidth}
    \begin{center}
      %\fbox{\rule{0pt}{2in} \rule{\linewidth}{0pt}}
      \includegraphics[width=\linewidth]{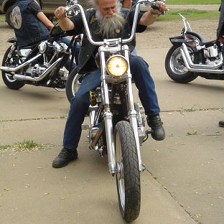}
      %\caption{}
    \end{center}
  \end{subfigure}\hfill\begin{subfigure}{0.095\linewidth}
  \begin{center}
    %\fbox{\rule{0pt}{2in} \rule{\linewidth}{0pt}}
    \includegraphics[width=\linewidth]{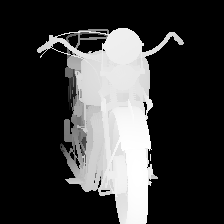}
    %\caption{}
  \end{center}
  \end{subfigure}\hfill\begin{subfigure}{0.095\linewidth}
  \begin{center}
    %\fbox{\rule{0pt}{2in} \rule{\linewidth}{0pt}}
    \includegraphics[width=\linewidth]{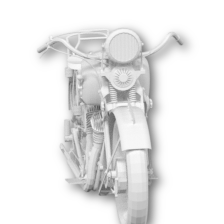}
    %\caption{}
  \end{center}
  \end{subfigure}\hfill\begin{subfigure}{0.095\linewidth}
  \begin{center}
    %\fbox{\rule{0pt}{2in} \rule{\linewidth}{0pt}}
    \includegraphics[width=\linewidth]{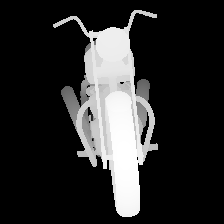}
    %\caption{}
  \end{center}
  \end{subfigure}\hfill\begin{subfigure}{0.095\linewidth}
  \begin{center}
    %\fbox{\rule{0pt}{2in} \rule{\linewidth}{0pt}}
    \includegraphics[width=\linewidth]{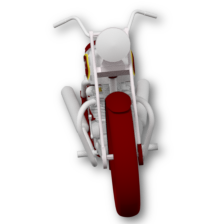}
    %\caption{}
  \end{center}
  \end{subfigure}\hfill\begin{subfigure}{0.095\linewidth}
  \begin{center}
    %\fbox{\rule{0pt}{2in} \rule{\linewidth}{0pt}}
    \includegraphics[width=\linewidth]{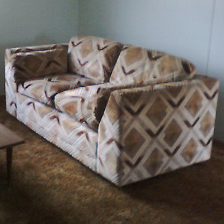}
    %\caption{}
  \end{center}
  \end{subfigure}\hfill\begin{subfigure}{0.095\linewidth}
  \begin{center}
    %\fbox{\rule{0pt}{2in} \rule{\linewidth}{0pt}}
    \includegraphics[width=\linewidth]{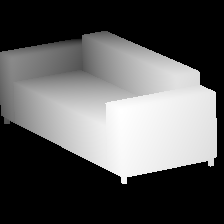}
    %\caption{}
  \end{center}
  \end{subfigure}\hfill\begin{subfigure}{0.095\linewidth}
  \begin{center}
    %\fbox{\rule{0pt}{2in} \rule{\linewidth}{0pt}}
    \includegraphics[width=\linewidth]{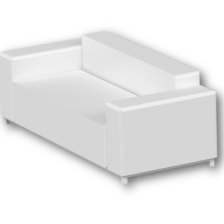}
    %\caption{}
  \end{center}
  \end{subfigure}\hfill\begin{subfigure}{0.095\linewidth}
  \begin{center}
    %\fbox{\rule{0pt}{2in} \rule{\linewidth}{0pt}}
    \includegraphics[width=\linewidth]{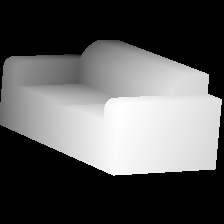}
    %\caption{}
  \end{center}
  \end{subfigure}\hfill\begin{subfigure}{0.095\linewidth}
  \begin{center}
    %\fbox{\rule{0pt}{2in} \rule{\linewidth}{0pt}}
    \includegraphics[width=\linewidth]{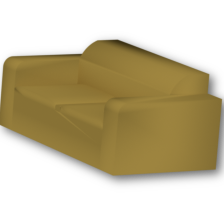}
    %\caption{}
  \end{center}
  \end{subfigure}\\[0.06cm]
  \begin{subfigure}{0.095\linewidth}
    \begin{center}
      %\fbox{\rule{0pt}{2in} \rule{\linewidth}{0pt}}
      \includegraphics[width=\linewidth]{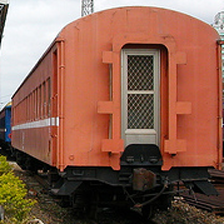}
      %\caption{}
    \end{center}
  \end{subfigure}\hfill\begin{subfigure}{0.095\linewidth}
  \begin{center}
    %\fbox{\rule{0pt}{2in} \rule{\linewidth}{0pt}}
    \includegraphics[width=\linewidth]{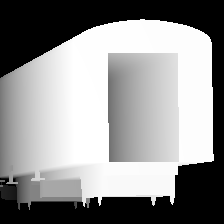}
    %\caption{}
  \end{center}
  \end{subfigure}\hfill\begin{subfigure}{0.095\linewidth}
  \begin{center}
    %\fbox{\rule{0pt}{2in} \rule{\linewidth}{0pt}}
    \includegraphics[width=\linewidth]{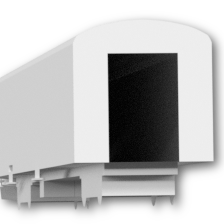}
    %\caption{}
  \end{center}
  \end{subfigure}\hfill\begin{subfigure}{0.095\linewidth}
  \begin{center}
    %\fbox{\rule{0pt}{2in} \rule{\linewidth}{0pt}}
    \includegraphics[width=\linewidth]{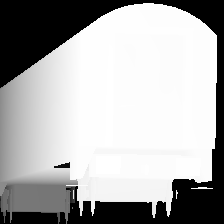}
    %\caption{}
  \end{center}
  \end{subfigure}\hfill\begin{subfigure}{0.095\linewidth}
  \begin{center}
    %\fbox{\rule{0pt}{2in} \rule{\linewidth}{0pt}}
    \includegraphics[width=\linewidth]{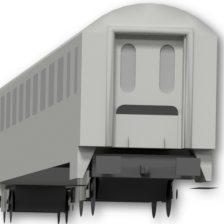}
    %\caption{}
  \end{center}
  \end{subfigure}\hfill\begin{subfigure}{0.095\linewidth}
  \begin{center}
    %\fbox{\rule{0pt}{2in} \rule{\linewidth}{0pt}}
    \includegraphics[width=\linewidth]{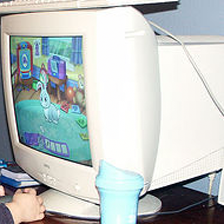}
    %\caption{}
  \end{center}
  \end{subfigure}\hfill\begin{subfigure}{0.095\linewidth}
  \begin{center}
    %\fbox{\rule{0pt}{2in} \rule{\linewidth}{0pt}}
    \includegraphics[width=\linewidth]{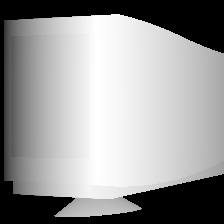}
    %\caption{}
  \end{center}
  \end{subfigure}\hfill\begin{subfigure}{0.095\linewidth}
  \begin{center}
    %\fbox{\rule{0pt}{2in} \rule{\linewidth}{0pt}}
    \includegraphics[width=\linewidth]{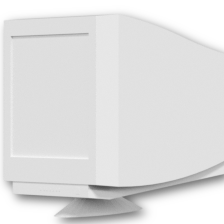}
    %\caption{}
  \end{center}
  \end{subfigure}\hfill\begin{subfigure}{0.095\linewidth}
  \begin{center}
    %\fbox{\rule{0pt}{2in} \rule{\linewidth}{0pt}}
    \includegraphics[width=\linewidth]{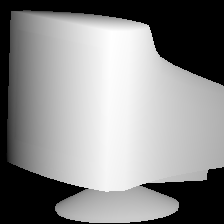}
    %\caption{}
  \end{center}
  \end{subfigure}\hfill\begin{subfigure}{0.095\linewidth}
  \begin{center}
    %\fbox{\rule{0pt}{2in} \rule{\linewidth}{0pt}}
    \includegraphics[width=\linewidth]{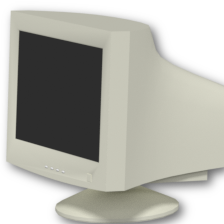}
    %\caption{}
  \end{center}
  \end{subfigure}
  \caption{Qualitative results for 3D pose estimation and 3D model retrieval from ShapeNet given images from Pascal3D+ for all twelve categories. For each category, we show: the query RGB image; the depth image and RGB rendering of the ground truth 3D model from Pascal3D+ under the ground truth pose from Pascal3D+; the depth image and RGB rendering of our retrieved 3D model from ShapeNet under our predicted pose.}
  \label{fig:retrieval_supp}
\end{figure*}

Fig.~\ref{fig:retrieval_supp} shows additional qualitative results for 3D model retrieval from ShapeNet~\cite{shapenet2015} given previously unseen images from Pascal3D+~\cite{xiang2014beyond} validation data for all twelve categories. Our approach predicts accurate 3D poses and 3D models for objects of different categories.

\begin{figure*}
\begin{center}
\begin{subfigure}{0.93\linewidth} \begin{subfigure}{0.16\linewidth} 
\begin{center}
    %\fbox{\rule{0pt}{2in} \rule{\linewidth}{0pt}}
    \includegraphics[width=\linewidth]{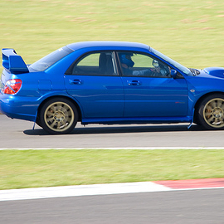}
    %\caption{}
\end{center} \end{subfigure}\hfill\begin{subfigure}{0.16\linewidth}
\begin{center}
	%\fbox{\rule{0pt}{2in} \rule{\linewidth}{0pt}}
	\includegraphics[width=\linewidth]{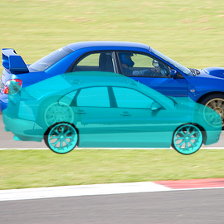}
	%\caption{}
\end{center} 
\end{subfigure}\hfill\begin{subfigure}{0.16\linewidth}
\begin{center}
    %\fbox{\rule{0pt}{2in} \rule{\linewidth}{0pt}}
    \includegraphics[width=\linewidth]{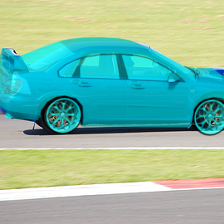}
    %\caption{}
\end{center} \end{subfigure}\hfill\begin{subfigure}{0.16\linewidth}
\begin{center}
    %\fbox{\rule{0pt}{2in} \rule{\linewidth}{0pt}}
    \includegraphics[width=\linewidth]{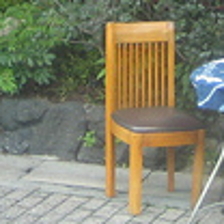}
    %\caption{}
\end{center} \end{subfigure}\hfill\begin{subfigure}{0.16\linewidth}
\begin{center}
    %\fbox{\rule{0pt}{2in} \rule{\linewidth}{0pt}}
    \includegraphics[width=\linewidth]{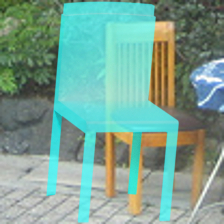}
    %\caption{}
\end{center} \end{subfigure}\hfill\begin{subfigure}{0.16\linewidth}
\begin{center}
    %\fbox{\rule{0pt}{2in} \rule{\linewidth}{0pt}}
    \includegraphics[width=\linewidth]{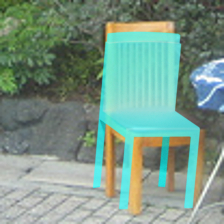}
    %\caption{}
\end{center} \end{subfigure}\\[0.1cm] \begin{subfigure}{0.16\linewidth}
\begin{center}
	%\fbox{\rule{0pt}{2in} \rule{\linewidth}{0pt}}
	\includegraphics[width=\linewidth]{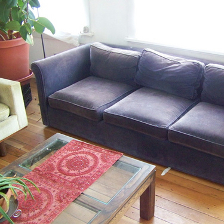}
	%\caption{}
\end{center} \end{subfigure}\hfill\begin{subfigure}{0.16\linewidth}
\begin{center}
	%\fbox{\rule{0pt}{2in} \rule{\linewidth}{0pt}}
	\includegraphics[width=\linewidth]{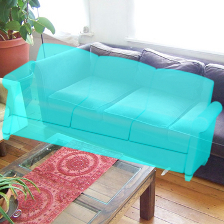}
	%\caption{}
\end{center} \end{subfigure}\hfill\begin{subfigure}{0.16\linewidth}
\begin{center}
	%\fbox{\rule{0pt}{2in} \rule{\linewidth}{0pt}}
	\includegraphics[width=\linewidth]{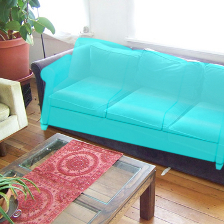}
	%\caption{}
\end{center} \end{subfigure}\hfill\begin{subfigure}{0.16\linewidth}
\begin{center}
%\fbox{\rule{0pt}{2in} \rule{\linewidth}{0pt}}
\includegraphics[width=\linewidth]{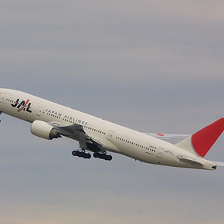}
%\caption{}
\end{center} \end{subfigure}\hfill\begin{subfigure}{0.16\linewidth}
\begin{center}
%\fbox{\rule{0pt}{2in} \rule{\linewidth}{0pt}}
\includegraphics[width=\linewidth]{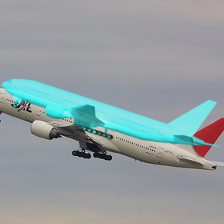}
%\caption{}
\end{center} \end{subfigure}\hfill\begin{subfigure}{0.16\linewidth}
\begin{center}
%\fbox{\rule{0pt}{2in} \rule{\linewidth}{0pt}}
\includegraphics[width=\linewidth]{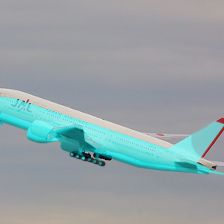}
%\caption{}
\end{center} \end{subfigure}\\[0.1cm] \begin{subfigure}{0.16\linewidth}
\begin{center}
%\fbox{\rule{0pt}{2in} \rule{\linewidth}{0pt}}
\includegraphics[width=\linewidth]{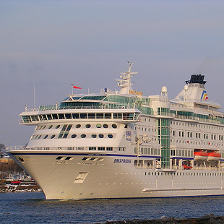}
%\caption{}
\end{center} \end{subfigure}\hfill\begin{subfigure}{0.16\linewidth}
\begin{center}
%\fbox{\rule{0pt}{2in} \rule{\linewidth}{0pt}}
\includegraphics[width=\linewidth]{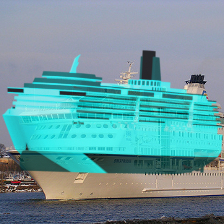}
%\caption{}
\end{center} \end{subfigure}\hfill\begin{subfigure}{0.16\linewidth}
\begin{center}
%\fbox{\rule{0pt}{2in} \rule{\linewidth}{0pt}}
\includegraphics[width=\linewidth]{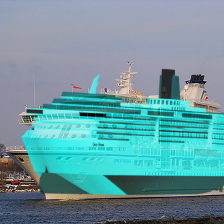}
%\caption{}
\end{center} \end{subfigure}\hfill\begin{subfigure}{0.16\linewidth}
\begin{center}
%\fbox{\rule{0pt}{2in} \rule{\linewidth}{0pt}}
\includegraphics[width=\linewidth]{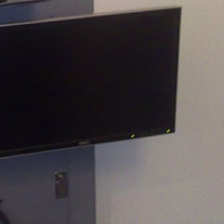}
%\caption{}
\end{center} \end{subfigure}\hfill\begin{subfigure}{0.16\linewidth}
\begin{center}
%\fbox{\rule{0pt}{2in} \rule{\linewidth}{0pt}}
\includegraphics[width=\linewidth]{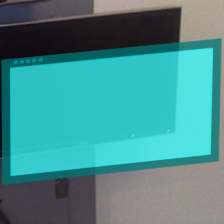}
%\caption{}
\end{center} \end{subfigure}\hfill\begin{subfigure}{0.16\linewidth}
\begin{center}
%\fbox{\rule{0pt}{2in} \rule{\linewidth}{0pt}}
\includegraphics[width=\linewidth]{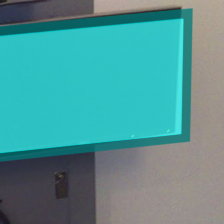}
%\caption{}
\end{center} \end{subfigure}\end{subfigure}\end{center}\vspace*{-0.3cm}\caption{We use our predicted 6-DoF pose and 3-DoF dimensions to refine the alignment between the object and a
rendering. Left: A detected object, which is not centered on the image
window. Middle: A rendering which just uses our predicted 3-DoF rotation. Right:
A rendering which uses our predicted 6-DoF pose and 3-DoF dimensions.}
    \label{fig:alignment_supp}
\end{figure*}

Fig.~\ref{fig:alignment_supp} presents further 3D model alignment results for object detections which are not fully accurate. We significantly improve the alignment between the object in the image and an RGB rendering of our retrieved 3D model by taking advantage of our predicted 6-DoF pose and 3-DoF dimensions compared to just using a 3-DoF viewpoint.

\section{Failure Modes}
\label{sec:fc}

Most failure cases of our 3D pose estimation on Pascal3D+ relate to low-resolution or ambiguous objects.

\begin{figure}
	\begin{subfigure}{0.19\linewidth}
		\begin{center}
			%\fbox{\rule{0pt}{2in} \rule{\linewidth}{0pt}}
			\includegraphics[width=\linewidth]{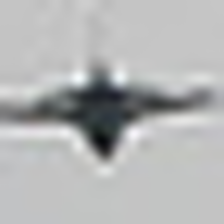}
			%\caption{}
		\end{center}
	\end{subfigure}\hfill\begin{subfigure}{0.19\linewidth}
		\begin{center}
			%\fbox{\rule{0pt}{2in} \rule{\linewidth}{0pt}}
			\includegraphics[width=\linewidth]{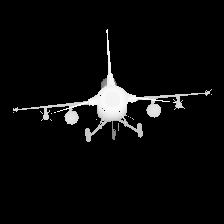}
			%\caption{}
		\end{center}
	\end{subfigure}\hfill\begin{subfigure}{0.19\linewidth}
		\begin{center}
			%\fbox{\rule{0pt}{2in} \rule{\linewidth}{0pt}}
			\includegraphics[width=\linewidth]{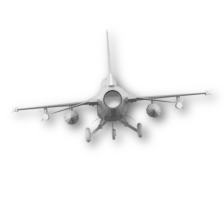}
			%\caption{}
		\end{center}
	\end{subfigure}\hfill\begin{subfigure}{0.19\linewidth}
		\begin{center}
			%\fbox{\rule{0pt}{2in} \rule{\linewidth}{0pt}}
			\includegraphics[width=\linewidth]{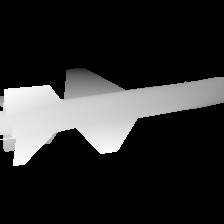}
			%\caption{}
		\end{center}
	\end{subfigure}\hfill\begin{subfigure}{0.19\linewidth}
		\begin{center}
			%\fbox{\rule{0pt}{2in} \rule{\linewidth}{0pt}}
			\includegraphics[width=\linewidth]{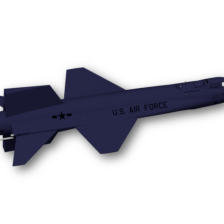}
			%\caption{}
		\end{center}
	\end{subfigure}\\[0.06cm]
\begin{subfigure}{0.19\linewidth}
		\begin{center}
			%\fbox{\rule{0pt}{2in} \rule{\linewidth}{0pt}}
			\includegraphics[width=\linewidth]{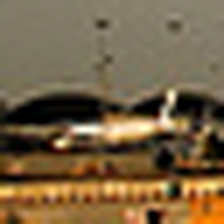}
			%\caption{}
		\end{center}
	\end{subfigure}\hfill\begin{subfigure}{0.19\linewidth}
		\begin{center}
			%\fbox{\rule{0pt}{2in} \rule{\linewidth}{0pt}}
			\includegraphics[width=\linewidth]{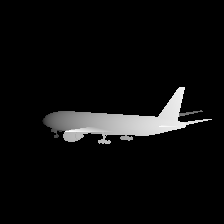}
			%\caption{}
		\end{center}
	\end{subfigure}\hfill\begin{subfigure}{0.19\linewidth}
		\begin{center}
			%\fbox{\rule{0pt}{2in} \rule{\linewidth}{0pt}}
			\includegraphics[width=\linewidth]{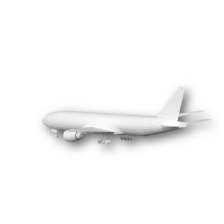}
			%\caption{}
		\end{center}
	\end{subfigure}\hfill\begin{subfigure}{0.19\linewidth}
		\begin{center}
			%\fbox{\rule{0pt}{2in} \rule{\linewidth}{0pt}}
			\includegraphics[width=\linewidth]{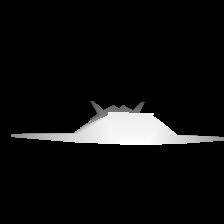}
			%\caption{}
		\end{center}
	\end{subfigure}\hfill\begin{subfigure}{0.19\linewidth}
		\begin{center}
			%\fbox{\rule{0pt}{2in} \rule{\linewidth}{0pt}}
			\includegraphics[width=\linewidth]{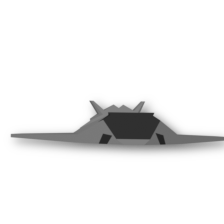}
			%\caption{}
		\end{center}
	\end{subfigure}\\[0.06cm]
	\begin{subfigure}{0.19\linewidth}
	\begin{center}
		%\fbox{\rule{0pt}{2in} \rule{\linewidth}{0pt}}
		\includegraphics[width=\linewidth]{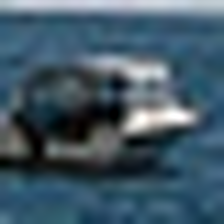}
		%\caption{}
	\end{center}
	\end{subfigure}\hfill\begin{subfigure}{0.19\linewidth}
	\begin{center}
		%\fbox{\rule{0pt}{2in} \rule{\linewidth}{0pt}}
		\includegraphics[width=\linewidth]{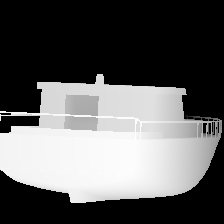}
		%\caption{}
	\end{center}
	\end{subfigure}\hfill\begin{subfigure}{0.19\linewidth}
	\begin{center}
		%\fbox{\rule{0pt}{2in} \rule{\linewidth}{0pt}}
		\includegraphics[width=\linewidth]{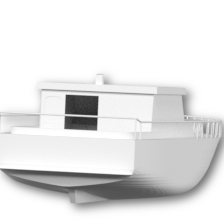}
		%\caption{}
	\end{center}
	\end{subfigure}\hfill\begin{subfigure}{0.19\linewidth}
	\begin{center}
		%\fbox{\rule{0pt}{2in} \rule{\linewidth}{0pt}}
		\includegraphics[width=\linewidth]{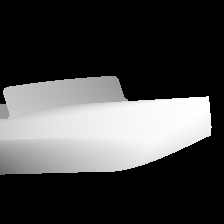}
		%\caption{}
	\end{center}
	\end{subfigure}\hfill\begin{subfigure}{0.19\linewidth}
	\begin{center}
		%\fbox{\rule{0pt}{2in} \rule{\linewidth}{0pt}}
		\includegraphics[width=\linewidth]{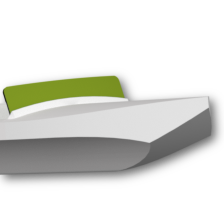}
		%\caption{}
	\end{center}
	\end{subfigure}\\[0.06cm]
	\begin{subfigure}{0.19\linewidth}
	\begin{center}
		%\fbox{\rule{0pt}{2in} \rule{\linewidth}{0pt}}
		\includegraphics[width=\linewidth]{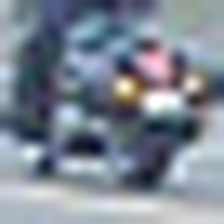}
		%\caption{}
	\end{center}
	\end{subfigure}\hfill\begin{subfigure}{0.19\linewidth}
	\begin{center}
		%\fbox{\rule{0pt}{2in} \rule{\linewidth}{0pt}}
		\includegraphics[width=\linewidth]{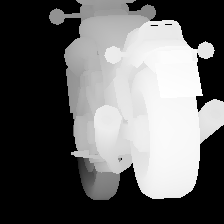}
		%\caption{}
	\end{center}
	\end{subfigure}\hfill\begin{subfigure}{0.19\linewidth}
	\begin{center}
		%\fbox{\rule{0pt}{2in} \rule{\linewidth}{0pt}}
		\includegraphics[width=\linewidth]{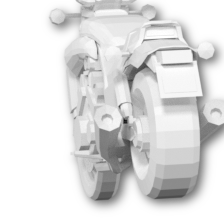}
		%\caption{}
	\end{center}
	\end{subfigure}\hfill\begin{subfigure}{0.19\linewidth}
	\begin{center}
		%\fbox{\rule{0pt}{2in} \rule{\linewidth}{0pt}}
		\includegraphics[width=\linewidth]{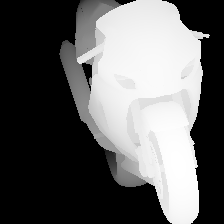}
		%\caption{}
	\end{center}
	\end{subfigure}\hfill\begin{subfigure}{0.19\linewidth}
	\begin{center}
		%\fbox{\rule{0pt}{2in} \rule{\linewidth}{0pt}}
		\includegraphics[width=\linewidth]{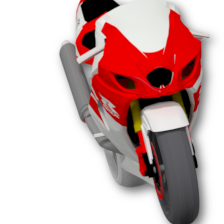}
		%\caption{}
	\end{center}
	\end{subfigure}
	\caption{3D pose estimation fails due to low-resolution image windows (same image arrangement as in Fig.~\ref{fig:retrieval_supp}). In fact, for more than 55\% of Pascal3D+ validation detections the longer edge of the 2D image window is smaller than 224 pixel, which is the fixed spatial input size of pre-trained CNNs like VGG~\cite{simonyan2014very} or ResNet~\cite{he2016deep,he2016identity}. If the resolution is too low, we cannot predict an accurate 3D pose.}
	\label{fig:failure_pose_blur}
\end{figure}

Fig.~\ref{fig:failure_pose_blur} shows 3D pose estimation results on low-resolution image windows from Pascal3D+ validation data.  After re-scaling, the over-smoothed input RGB images lack details and sharp discontinuities, which results in incorrect pose predictions.  In fact, even for a human it is difficult to identify the correct object poses in these examples.

\begin{figure}
	\begin{subfigure}{0.19\linewidth}
		\begin{center}
			%\fbox{\rule{0pt}{2in} \rule{\linewidth}{0pt}}
			\includegraphics[width=\linewidth]{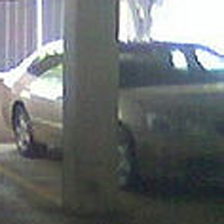}
			%\caption{}
		\end{center}
	\end{subfigure}\hfill\begin{subfigure}{0.19\linewidth}
		\begin{center}
			%\fbox{\rule{0pt}{2in} \rule{\linewidth}{0pt}}
			\includegraphics[width=\linewidth]{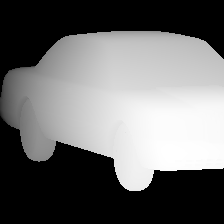}
			%\caption{}
		\end{center}
	\end{subfigure}\hfill\begin{subfigure}{0.19\linewidth}
		\begin{center}
			%\fbox{\rule{0pt}{2in} \rule{\linewidth}{0pt}}
			\includegraphics[width=\linewidth]{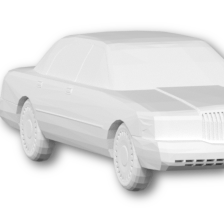}
			%\caption{}
		\end{center}
	\end{subfigure}\hfill\begin{subfigure}{0.19\linewidth}
		\begin{center}
			%\fbox{\rule{0pt}{2in} \rule{\linewidth}{0pt}}
			\includegraphics[width=\linewidth]{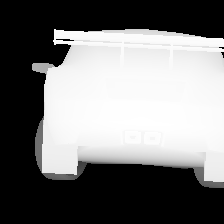}
			%\caption{}
		\end{center}
	\end{subfigure}\hfill\begin{subfigure}{0.19\linewidth}
		\begin{center}
			%\fbox{\rule{0pt}{2in} \rule{\linewidth}{0pt}}
			\includegraphics[width=\linewidth]{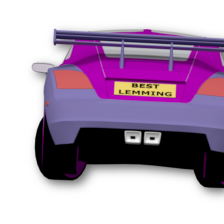}
			%\caption{}
		\end{center}
	\end{subfigure}\\[0.06cm]
	\begin{subfigure}{0.19\linewidth}
	\begin{center}
		%\fbox{\rule{0pt}{2in} \rule{\linewidth}{0pt}}
		\includegraphics[width=\linewidth]{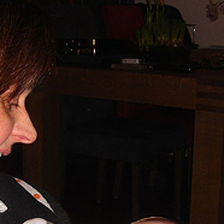}
		%\caption{}
	\end{center}
	\end{subfigure}\hfill\begin{subfigure}{0.19\linewidth}
	\begin{center}
		%\fbox{\rule{0pt}{2in} \rule{\linewidth}{0pt}}
		\includegraphics[width=\linewidth]{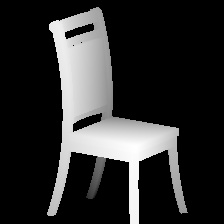}
		%\caption{}
	\end{center}
	\end{subfigure}\hfill\begin{subfigure}{0.19\linewidth}
	\begin{center}
		%\fbox{\rule{0pt}{2in} \rule{\linewidth}{0pt}}
		\includegraphics[width=\linewidth]{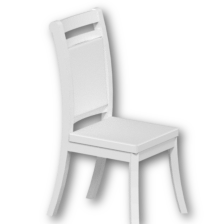}
		%\caption{}
	\end{center}
	\end{subfigure}\hfill\begin{subfigure}{0.19\linewidth}
	\begin{center}
		%\fbox{\rule{0pt}{2in} \rule{\linewidth}{0pt}}
		\includegraphics[width=\linewidth]{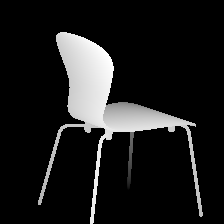}
		%\caption{}
	\end{center}
	\end{subfigure}\hfill\begin{subfigure}{0.19\linewidth}
	\begin{center}
		%\fbox{\rule{0pt}{2in} \rule{\linewidth}{0pt}}
		\includegraphics[width=\linewidth]{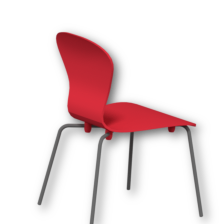}
		%\caption{}
	\end{center}
	\end{subfigure}\\[0.06cm]
	\begin{subfigure}{0.19\linewidth}
	\begin{center}
		%\fbox{\rule{0pt}{2in} \rule{\linewidth}{0pt}}
		\includegraphics[width=\linewidth]{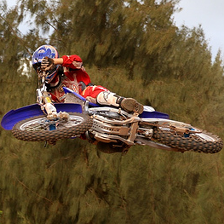}
		%\caption{}
	\end{center}
	\end{subfigure}\hfill\begin{subfigure}{0.19\linewidth}
	\begin{center}
		%\fbox{\rule{0pt}{2in} \rule{\linewidth}{0pt}}
		\includegraphics[width=\linewidth]{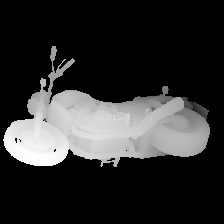}
		%\caption{}
	\end{center}
	\end{subfigure}\hfill\begin{subfigure}{0.19\linewidth}
	\begin{center}
		%\fbox{\rule{0pt}{2in} \rule{\linewidth}{0pt}}
		\includegraphics[width=\linewidth]{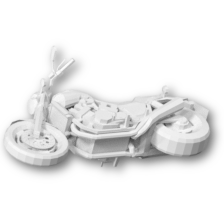}
		%\caption{}
	\end{center}
	\end{subfigure}\hfill\begin{subfigure}{0.19\linewidth}
	\begin{center}
		%\fbox{\rule{0pt}{2in} \rule{\linewidth}{0pt}}
		\includegraphics[width=\linewidth]{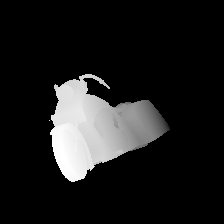}
		%\caption{}
	\end{center}
	\end{subfigure}\hfill\begin{subfigure}{0.19\linewidth}
	\begin{center}
		%\fbox{\rule{0pt}{2in} \rule{\linewidth}{0pt}}
		\includegraphics[width=\linewidth]{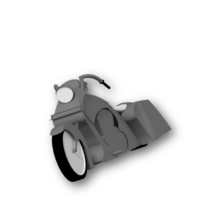}
		%\caption{}
	\end{center}
	\end{subfigure}\\[0.06cm]
\begin{subfigure}{0.19\linewidth}
\begin{center}
	%\fbox{\rule{0pt}{2in} \rule{\linewidth}{0pt}}
	\includegraphics[width=\linewidth]{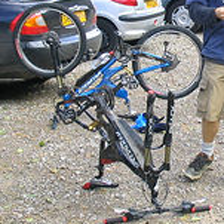}
	%\caption{}
\end{center}
\end{subfigure}\hfill\begin{subfigure}{0.19\linewidth}
\begin{center}
	%\fbox{\rule{0pt}{2in} \rule{\linewidth}{0pt}}
	\includegraphics[width=\linewidth]{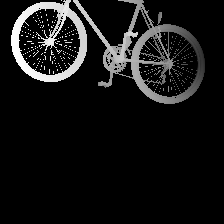}
	%\caption{}
\end{center}
\end{subfigure}\hfill\begin{subfigure}{0.19\linewidth}
\begin{center}
	%\fbox{\rule{0pt}{2in} \rule{\linewidth}{0pt}}
	\includegraphics[width=\linewidth]{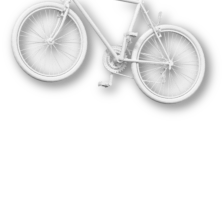}
	%\caption{}
\end{center}
\end{subfigure}\hfill\begin{subfigure}{0.19\linewidth}
\begin{center}
	%\fbox{\rule{0pt}{2in} \rule{\linewidth}{0pt}}
	\includegraphics[width=\linewidth]{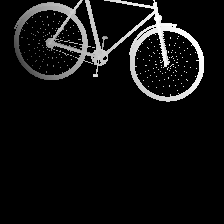}
	%\caption{}
\end{center}
\end{subfigure}\hfill\begin{subfigure}{0.19\linewidth}
\begin{center}
	%\fbox{\rule{0pt}{2in} \rule{\linewidth}{0pt}}
	\includegraphics[width=\linewidth]{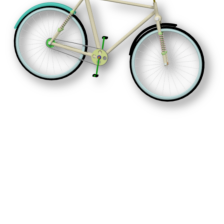}
	%\caption{}
\end{center}
\end{subfigure}
	\caption{3D pose estimation fails in difficult situations (same image arrangement as in Fig.~\ref{fig:retrieval_supp}). We observe that heavy occlusions (first row), bad illumination conditions (second row) and difficult object poses (third and fourth row), which are far from the poses seen during training, result in incorrect pose predictions. In the last row, we see that not even the annotated ground truth pose is correct.}
	\label{fig:failure_pose_difficult}
\end{figure}

Fig.~\ref{fig:failure_pose_difficult} shows additional failure cases, observing that heavy occlusions, bad illumination conditions and difficult object poses, which are far from the poses seen during training, result in incorrect pose predictions.

\begin{figure}
	\begin{subfigure}{0.23\linewidth}
		\begin{center}
			%\fbox{\rule{0pt}{2in} \rule{\linewidth}{0pt}}
			\includegraphics[width=\linewidth]{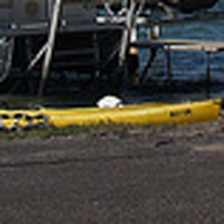}
			%\caption{}
		\end{center}
	\end{subfigure}\hfill\begin{subfigure}{0.23\linewidth}
		\begin{center}
			%\fbox{\rule{0pt}{2in} \rule{\linewidth}{0pt}}
			\includegraphics[width=\linewidth]{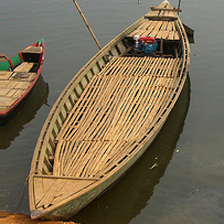}
			%\caption{}
		\end{center}
	\end{subfigure}\hfill\begin{subfigure}{0.23\linewidth}
		\begin{center}
			%\fbox{\rule{0pt}{2in} \rule{\linewidth}{0pt}}
			\includegraphics[width=\linewidth]{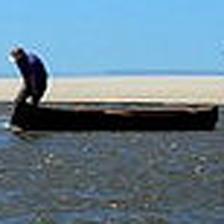}
			%\caption{}
		\end{center}
	\end{subfigure}\hfill\begin{subfigure}{0.23\linewidth}
		\begin{center}
			%\fbox{\rule{0pt}{2in} \rule{\linewidth}{0pt}}
			\includegraphics[width=\linewidth]{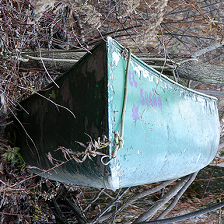}
			%\caption{}
		\end{center}
	\end{subfigure}\\[0.18cm]
\begin{subfigure}{0.23\linewidth}
		\begin{center}
			%\fbox{\rule{0pt}{2in} \rule{\linewidth}{0pt}}
			\includegraphics[width=\linewidth]{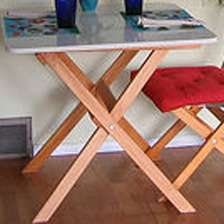}
			%\caption{}
		\end{center}
	\end{subfigure}\hfill\begin{subfigure}{0.23\linewidth}
		\begin{center}
			%\fbox{\rule{0pt}{2in} \rule{\linewidth}{0pt}}
			\includegraphics[width=\linewidth]{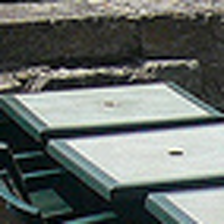}
			%\caption{}
		\end{center}
	\end{subfigure}\hfill\begin{subfigure}{0.23\linewidth}
		\begin{center}
			%\fbox{\rule{0pt}{2in} \rule{\linewidth}{0pt}}
			\includegraphics[width=\linewidth]{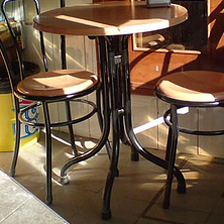}
			%\caption{}
		\end{center}
	\end{subfigure}\hfill\begin{subfigure}{0.23\linewidth}
		\begin{center}
			%\fbox{\rule{0pt}{2in} \rule{\linewidth}{0pt}}
			\includegraphics[width=\linewidth]{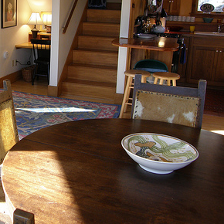}
			%\caption{}
		\end{center}
	\end{subfigure}
	\caption{Objects with ambiguous poses from Pascal3D+ validation data. First row: It is impossible to differentiate between the front and back of symmetric boats. Second row: Tables which are ambiguous with respect to an azimuth rotation of $\pi$ (first image), $\frac{\pi}{2}$ (second and third image) or even have an axis of symmetry (fourth image).}
	\label{fig:failure_pose_ambi}
\end{figure}

As shown  in Fig.~\ref{fig:failure_pose_ambi},  some objects from  Pascal3D+ are symmetrical,  which makes  their  poses not  well defined.   For  example, it  is impossible to differentiate  between the front and back of  a symmetric unmanned boat.  This issue  is even more  apparent for tables:  Many tables are  ambiguous with respect to an azimuth rotation of $\pi$, $\frac{\pi}{2}$ or even have an axis of symmetry, such as a round table. When our approach predicts  one of the possible poses that  is not the annotated ground  truth  pose, this  is  considered  as a  mistake  by  the commonly  used evaluation protocol~\cite{tulsiani2015viewpoints}.

\begin{figure}
	\begin{subfigure}{0.19\linewidth}
		\begin{center}
			%\fbox{\rule{0pt}{2in} \rule{\linewidth}{0pt}}
			\includegraphics[width=\linewidth]{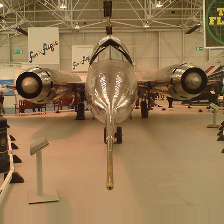}
			%\caption{}
		\end{center}
	\end{subfigure}\hfill\begin{subfigure}{0.19\linewidth}
		\begin{center}
			%\fbox{\rule{0pt}{2in} \rule{\linewidth}{0pt}}
			\includegraphics[width=\linewidth]{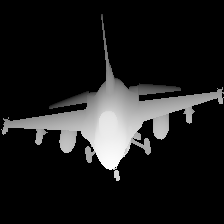}
			%\caption{}
		\end{center}
	\end{subfigure}\hfill\begin{subfigure}{0.19\linewidth}
		\begin{center}
			%\fbox{\rule{0pt}{2in} \rule{\linewidth}{0pt}}
			\includegraphics[width=\linewidth]{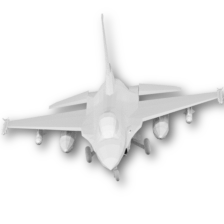}
			%\caption{}
		\end{center}
	\end{subfigure}\hfill\begin{subfigure}{0.19\linewidth}
		\begin{center}
			%\fbox{\rule{0pt}{2in} \rule{\linewidth}{0pt}}
			\includegraphics[width=\linewidth]{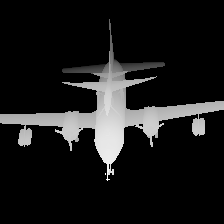}
			%\caption{}
		\end{center}
	\end{subfigure}\hfill\begin{subfigure}{0.19\linewidth}
		\begin{center}
			%\fbox{\rule{0pt}{2in} \rule{\linewidth}{0pt}}
			\includegraphics[width=\linewidth]{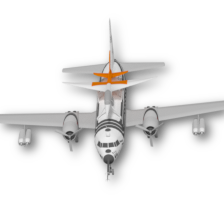}
			%\caption{}
		\end{center}
	\end{subfigure}\\[0.06cm]
\begin{subfigure}{0.19\linewidth}
		\begin{center}
			%\fbox{\rule{0pt}{2in} \rule{\linewidth}{0pt}}
			\includegraphics[width=\linewidth]{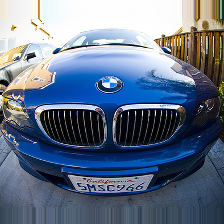}
			%\caption{}
		\end{center}
	\end{subfigure}\hfill\begin{subfigure}{0.19\linewidth}
		\begin{center}
			%\fbox{\rule{0pt}{2in} \rule{\linewidth}{0pt}}
			\includegraphics[width=\linewidth]{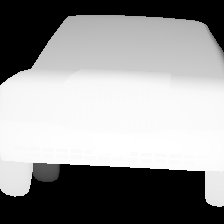}
			%\caption{}
		\end{center}
	\end{subfigure}\hfill\begin{subfigure}{0.19\linewidth}
		\begin{center}
			%\fbox{\rule{0pt}{2in} \rule{\linewidth}{0pt}}
			\includegraphics[width=\linewidth]{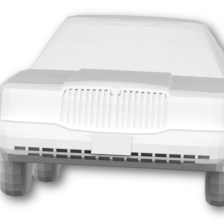}
			%\caption{}
		\end{center}
	\end{subfigure}\hfill\begin{subfigure}{0.19\linewidth}
		\begin{center}
			%\fbox{\rule{0pt}{2in} \rule{\linewidth}{0pt}}
			\includegraphics[width=\linewidth]{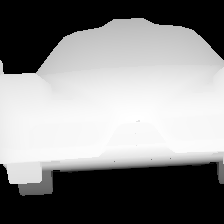}
			%\caption{}
		\end{center}
	\end{subfigure}\hfill\begin{subfigure}{0.19\linewidth}
		\begin{center}
			%\fbox{\rule{0pt}{2in} \rule{\linewidth}{0pt}}
			\includegraphics[width=\linewidth]{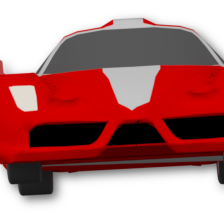}
			%\caption{}
		\end{center}
	\end{subfigure}\\[0.06cm]
	\begin{subfigure}{0.19\linewidth}
	\begin{center}
		%\fbox{\rule{0pt}{2in} \rule{\linewidth}{0pt}}
		\includegraphics[width=\linewidth]{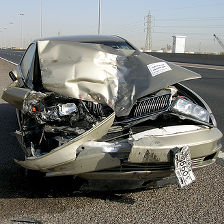}
		%\caption{}
	\end{center}
	\end{subfigure}\hfill\begin{subfigure}{0.19\linewidth}
	\begin{center}
		%\fbox{\rule{0pt}{2in} \rule{\linewidth}{0pt}}
		\includegraphics[width=\linewidth]{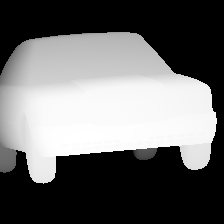}
		%\caption{}
	\end{center}
	\end{subfigure}\hfill\begin{subfigure}{0.19\linewidth}
	\begin{center}
		%\fbox{\rule{0pt}{2in} \rule{\linewidth}{0pt}}
		\includegraphics[width=\linewidth]{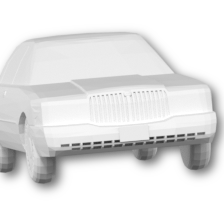}
		%\caption{}
	\end{center}
	\end{subfigure}\hfill\begin{subfigure}{0.19\linewidth}
	\begin{center}
		%\fbox{\rule{0pt}{2in} \rule{\linewidth}{0pt}}
		\includegraphics[width=\linewidth]{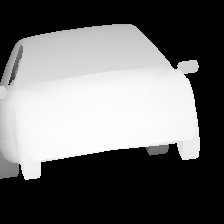}
		%\caption{}
	\end{center}
	\end{subfigure}\hfill\begin{subfigure}{0.19\linewidth}
	\begin{center}
		%\fbox{\rule{0pt}{2in} \rule{\linewidth}{0pt}}
		\includegraphics[width=\linewidth]{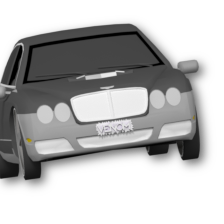}
		%\caption{}
	\end{center}
	\end{subfigure}\\[0.06cm]
	\begin{subfigure}{0.19\linewidth}
	\begin{center}
		%\fbox{\rule{0pt}{2in} \rule{\linewidth}{0pt}}
		\includegraphics[width=\linewidth]{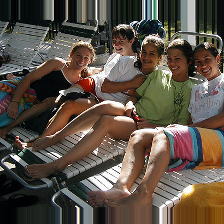}
		%\caption{}
	\end{center}
	\end{subfigure}\hfill\begin{subfigure}{0.19\linewidth}
	\begin{center}
		%\fbox{\rule{0pt}{2in} \rule{\linewidth}{0pt}}
		\includegraphics[width=\linewidth]{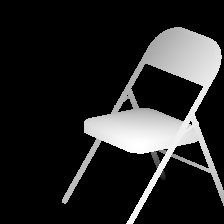}
		%\caption{}
	\end{center}
	\end{subfigure}\hfill\begin{subfigure}{0.19\linewidth}
	\begin{center}
		%\fbox{\rule{0pt}{2in} \rule{\linewidth}{0pt}}
		\includegraphics[width=\linewidth]{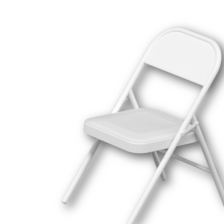}
		%\caption{}
	\end{center}
	\end{subfigure}\hfill\begin{subfigure}{0.19\linewidth}
	\begin{center}
		%\fbox{\rule{0pt}{2in} \rule{\linewidth}{0pt}}
		\includegraphics[width=\linewidth]{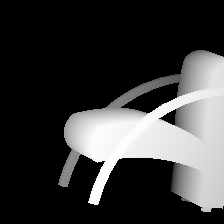}
		%\caption{}
	\end{center}
	\end{subfigure}\hfill\begin{subfigure}{0.19\linewidth}
	\begin{center}
		%\fbox{\rule{0pt}{2in} \rule{\linewidth}{0pt}}
		\includegraphics[width=\linewidth]{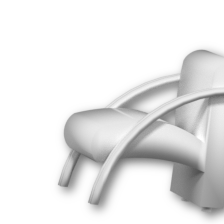}
		%\caption{}
	\end{center}
	\end{subfigure}
	\caption{3D  model retrieval  results for  challenging cases  where pose
          estimation   was   successful   (same    image   arrangement   as   in
          Fig.~\ref{fig:retrieval_supp}).  The test  images can exhibit fish-eye
          effects  due to  wide-angle  lenses (first  and  second row),  contain
          deformed or  demolished objects  (third row),  or objects  under heavy
          occlusions (fourth row), which  disturb object retrieval. Note however
          that the ground truth 3D models are not accurate.}
	\label{fig:failure_retrieval_difficult}
\end{figure}

Fig.~\ref{fig:failure_retrieval_difficult} shows that visual distortions due to wide-angle lenses (\ie,  fish-eye effects), deformed and demolished objects and heavy occlusions can disturb the model retrieval step, even if the pose estimation was successful.

\section{Implementation Details}
\label{sec:details}

In the following, we provide implementation details and other parameters used in our work:

\emph{Intrinsic camera parameters}: In Pascal3D+, the ground truth poses were computed from 2D-3D correspondences assuming the same intrinsic parameters for all images. We employ the same parameters in our approach.

\emph{Data augmentation}: Like others~\cite{mousavian20163d,pavlakos17object3d,su2015render,tulsiani2015viewpoints}, we perform data augmentation by jittering ground truth detections and exclude detections marked as occluded or truncated from the evaluation. Additionally, we augment samples for which the longer edge of the ground truth image window is greater than 224 pixel by applying Gaussian blurring with various kernel sizes and $\sigma$. We randomly sample negative example 3D models from the available data. All augmentation parameters are randomized after each training epoch.

\emph{Meta parameters}: We normalize the projections so that the image pixel range is mapped to the interval [0,1] and use the same Huber loss ($\delta = 0.01$) for all 19 estimated values. Experimentally, we found $\alpha=1$, $\beta=1e^{-5}$ and $\gamma=1e^{-3}$ to work well and set $m=1$.

\emph{Network parameters}: We use a batch size of 50, train our networks for 100 epochs and decrease the initial learning rate of $1e^{-4}$ by one order of magnitude after 50 and 90 epochs, and employ the Adam optimization algorithm.

\emph{3D dimensions}: For both Pascal3D+ and ShapeNet, 3D models are normalized to fit within a unit cube centered at the origin. Thus, we estimate 3D dimensions in model space in the range [0,1]. Since these dimensions tend to be consistent within a category, estimating them is not a major issue. Table~\ref{table:dims} shows quantitative results for 3D dimension estimation. We achieve high accuracy across all categories.

\begin{table}[!h]
	\footnotesize
	\centering
	\begin{tabular}{lccc}
		\toprule
		&x&y&z\\
		\midrule
		\textit{Median Absolute Error}&0.022&0.015&0.014\\
		\bottomrule\\[-0.2cm]
	\end{tabular}
	\caption{3D dimension estimation errors on Pascal3D+. We report the mean performance across all categories.}
	\label{table:dims}
\end{table}

\clearpage

\end{document}